\documentclass[10pt,journal,compsoc]{IEEEtran}
\ifCLASSOPTIONcompsoc
  \usepackage[nocompress]{cite}
\else
  \usepackage{cite}
\fi
\ifCLASSINFOpdf
\else
\fi
\usepackage{amsmath}
\usepackage{algorithm, algorithmicx}

\usepackage{array}

\ifCLASSOPTIONcompsoc
  \usepackage[caption=false,font=footnotesize,labelfont=sf,textfont=sf,subrefformat=parens,labelformat=parens]{subfig}
\else
  \usepackage[caption=false,font=footnotesize]{subfig}
\fi
\usepackage{fixltx2e}

\usepackage{bm,pifont}
\usepackage{verbatim}
\usepackage{graphicx}
\usepackage{booktabs}
\usepackage{multirow}
\usepackage{amssymb,amsfonts,nccmath, mathtools, amsthm}
\usepackage{booktabs}
\usepackage{footmisc}
\usepackage[noend]{algpseudocode}
\usepackage{upgreek}
\usepackage{balance}

\newtheorem{assumption}{Assumption}
\newtheorem{lemma}{Lemma}
\newtheorem{definition}{Definition}
\newtheorem{theorem}{Theorem}

\begin{document}
%
\title{Flexible Clustered Federated Learning for Client-Level Data Distribution Shift
}

\author{Moming~Duan,
	Duo~Liu,
	Xinyuan~Ji,
	Yu~Wu,
	Liang~Liang,
	Xianzhang~Chen,
	Yujuan~Tan
\IEEEcompsocitemizethanks{
	\IEEEcompsocthanksitem A preliminary version of this paper was accept at the IEEE 2021 International Symposium on Parallel and Distributed Processing with Applications(ISPA)~\cite{duan2020fedgroup}.~This longer version provides the convergence proof and extends previous analyse to the new setting in Section~\ref{sec:evaluation}.
	\IEEEcompsocthanksitem Moming~Duan, Duo~Liu, Xianzhang~Chen, Renping Liu, Yujuan~Tan are with Key Laboratory of Dependable Service Computing in Cyber Physical Society (Chongqing University), Ministry of Education, China, and College of Computer Science, Chongqing University, Chongqing 400044, P. R. China.
	\IEEEcompsocthanksitem Liang Liang is with School of Microelectronics and Communication Engineering, Chongqing University, Chongqing 400044, P.R.China
	}
}

%
%

\markboth{IEEE TRANSACTIONS ON PARALLEL AND DISTRIBUTED SYSTEMS,~Vol.~x, No.~x, August~x}%
{xxx \MakeLowercase{\textit{et al.}}: Computer Society Journals}
%



\IEEEtitleabstractindextext{%
\begin{abstract}
Federated Learning (FL) enables the multiple participating devices to collaboratively contribute to a global neural network model while keeping the training data locally. 
Unlike the centralized training setting, the non-IID, imbalanced (statistical heterogeneity) and distribution shifted training data of FL is distributed in the federated network, which will increase the divergences between the local models and the global model, further degrading performance. 
In this paper, we propose a flexible clustered federated learning (CFL) framework named FlexCFL, in which we 
1) group the training of clients based on the similarities between the clients' optimization directions for lower training divergence; 
2) implement an efficient newcomer device cold start mechanism for framework scalability and practicality;
3) flexibly migrate clients to meet the challenge of client-level data distribution shift.
FlexCFL can achieve improvements by dividing joint optimization into groups of sub-optimization and can strike a balance between accuracy and communication efficiency in the distribution shift environment.
The convergence and complexity are analyzed to demonstrate the efficiency of FlexCFL.
We also evaluate FlexCFL on several open datasets and made comparisons with related CFL frameworks. The results show that FlexCFL can significantly improve absolute test accuracy by $+10.6\%$ on FEMNIST compared to \textit{FedAvg}, $+3.5\%$ on FashionMNIST compared to \textit{FedProx}, $+8.4\%$ on MNIST compared to \textit{FeSEM}. The experiment results show that FlexCFL is also communication efficient in the distribution shift environment.

\end{abstract}

}

\maketitle

\IEEEdisplaynontitleabstractindextext

%
\IEEEpeerreviewmaketitle


\IEEEraisesectionheading{\section{Introduction}\label{sec:introduction}}

\IEEEPARstart{F}ederated Learning (FL)~\cite{konevcny2016federated,mcmahan2017communication,bonawitz2019towards,yang2019federated,li2020federated} is a promising distributed neural network training approach, which enables multiple end-users to collaboratively train a shared neural network model while keeping the training data decentralized. 
In practice, a FL server first distributes the global model to a random subset of participating clients (e.g. mobile and IoT devices). 
Then each client optimizes its local model by gradient descent based on its local data in parallel. 
Finally, the FL server averages all local models' updates or parameters and aggregates them to construct a new global model.
Unlike the traditional cloud-centric learning paradigm and the distributed machine learning frameworks based on Parameter Server~\cite{li2014scaling}, there is no need to transfer private data over the communication network during the FL training.
With the advantage of privacy-preserving, Federated Learning is currently the most attractive distributed machine learning framework.
Nevertheless, due to the FL server does not have the authority to access the user data or collect statistical information, some data preparation operations such as balancing and outlier detection be restricted. 
Therefore, the high statistical heterogeneity is a challenging problem in federated learning~\cite{li2020federated}.

To tackle heterogeneity in federated learning, several efforts have been made.
McMahan \textit{et al.} propose the vanilla FL framework \textit{Federated Averaging }(FedAvg)~\cite{mcmahan2017communication} and experimentally demonstrate that FedAvg is communication-efficient and can converge under statistical heterogeneity setting (non-IID). 
However, Zhao \textit{et al.}~\cite{zhao2018federated} show that the accuracy reduces $\sim$55\% for CNN trained on highly skewed CIFAR-10~\cite{krizhevsky2009learning}. 
The experiments based on VGG11~\cite{Simonyan15} by Sattler \textit{et al.}~\cite{sattler2019robust} show that non-IID data not only leads to accuracy degradation, but also reduces the convergence speed. 
Li \textit{et al.}\cite{Li2020On} theoretically analyze the convergence of FedAvg and indicates that the heterogeneity of data slows down the convergence for the strongly convex and smooth problem. 
In addition, Duan \textit{et al.}~\cite{duan2019astraea} demonstrate that global imbalanced data also has adverse effects on federated training. 
Unfortunately, the retrieval of model accuracy decreases as the local model diverges in~\cite{zhao2018federated,sattler2019robust,duan2019astraea}. 
Recently, Sattler \textit{et al.}~\cite{sattler2020clustered, sattler2020byzantine} propose a novel federated multi-task learning framework Cluster Federated Learning (CFL), which exploits geometric properties of FL loss surface to cluster the learning processes of clients based on their optimization direction, provides a new way of thinking about the statistical heterogeneity challenge. Many researchers follow up on CFL-based framework \cite{xie2020multi, ghosh2020efficient, briggs2020federated} and confirm CFL is more accurate than traditional FL with a consensus global model. However, the above CFL-based frameworks are inefficient in the large-scale federated training systems and ignore the presence of newcomer devices. 
Moreover, the data distribution shifts may degrade the clustering performance and are not involved in their experiments, which makes their empirical analysis insufficient.

In this paper, we present an efficient and flexible clustered federated learning framework FlexCFL.
To improve the efficiency, we leverage a novel decomposed data-driven measure called Euclidean distance of Decomposed Cosine similarity (EDC) for client clustering.
The main advantage of EDC is it can avoid the concentration phenomenon of $\ell_p$ distances in high dimensional data clustering~\cite{sarkar2019perfect}.
Another design that makes FlexCFL more practical is we maintain an auxiliary server to address the cold start issue of new devices.
Furthermore, FlexCFL can detect the client-level data distribution shift based on Wasserstein distance and migrate clients with affordable communication.

With the above methods, FlexCFL can significantly improve test accuracy by $+8.4\%$ on MNIST~\cite{lecun1998gradient}, $+40.9\%$ on FEMNIST~\cite{cohen2017emnist}, $+11.3\%$ on FashionMNIST~\cite{xiao2017fashion} compared to FedSEM. We show that FlexCFL has superior performance than FedAvg, FedProx~\cite{li2018federated} and FeSEM~\cite{xie2020multi}. Although FlexCFL achieves performance improvements similar to IFCA~\cite{ghosh2020efficient}, the latter has more communication consumption. The ablation studies of FlexCFL are provided to demonstrate the usefulness of our clustering and client migration strategies.

The main contributions of this paper are summarized as follows.
\begin{itemize}
	\item We propose a novel clustered federated learning framework, FlexCFL, and show its superiority on four open datasets (with statistical heterogeneity) compared to several FL and CFL frameworks.
	\item Our framework presents an efficient cold start strategy for the groups and the newcomers, which provides a new approach to improve the scalability and practicality of the previous works. We demonstrate its efficiency experimentally.
	\item We explore the impact of client-level data distribution shift on the clustered federated training and propose a communication-efficient client migration algorithm for FlexCFL. To the best of our knowledge, this is the first CFL framework considering the distribution shift challenge. In addition, we open source the code of FlexCFL to contribute to the community.
\end{itemize}

The rest of this paper is organized as follows.
Section~\ref{sec:background} provides the background of FL and an overview of related works.
Section~\ref{sec:design} shows the motivation and the design of FlexCFL. 
The convergence guarantee for FlexCFL is derived in Section~\ref{sec:convergence} and the evaluation results are presented and analyzed in Section~\ref{sec:evaluation}. 
Section~\ref{sec:conclusion} concludes the paper.


\section{background and related work}\label{sec:background}
\subsection{Federated Learning}
In this section, we introduce the most widely adopted FL algorithm FedAvg.
McMahan \textit{et al.} first introduce federated learning~\cite{mcmahan2017communication} and the vanilla FL optimization method FedAvg, which is designed to provide privacy-preserving support for distributed machine learning model training.
The distributed objective of FL is:

\begin{equation}
	\min\limits_{{\bm{w}}} \Big\{f(\bm{w}) \triangleq \sum_{k=1}^{N}p_k F_k(\bm{w})\Big\},
\end{equation}

Where $N$ is the number of clients, $p_k$ is the weight of the $k$-th device, $p_k \geqslant 0$, $\sum_{k}^{} p_k = 1$.
In statistical heterogeneity setting, the local objectives $F_k(\bm{w})$ measure the local expirical risk over possibly differing data distributions $p_{data}^{(k)}$. 

For a machine learning problem, we can set $F_k(\bm{w})$ to the user-specified loss function $L \big( \cdot;\cdot \big)$ of the predictions on examples $(\bm{x},\bm{y})$ made with model parameters $w$. Hence, the local objective is defined by

\begin{equation}
	F_k(\bm{w}) \triangleq \mathbb{E}_{(\bm{x},\bm{y})\sim{p}_{data}^{(k)}}L\big( \bm{x},\bm{y}; \bm{w} \big).
\end{equation}

The global objective $f(\bm{w})$ can be regarded as a joint objective function for multiple clients, and FL tries to minimize it by optimizing all local optimization objectives.

In practice, a typical implementation of FL system~\cite{bonawitz2019towards} includes a cloud server that maintains a global model and multiple participating devices, which communicate with the server through a network.
At each communication round $t$, the server selects a random subset $K$ of the active devices (i.e. clients) to participate in this round of training.
The server then broadcasts the latest global model $w^t$ to these clients for further local optimization.
Then each participating client optimizes the local objective function based on the device's data by its local solvers (e.g. SGD) with several local epochs $E$ in parallel.
Then, the locally-computed parameter updates $\Delta \bm{w}_i^t$ from these clients that completed the training within the time budget are collected and aggregated.
Finally, the server will update the global model to $w^{t+1}$ and finish the current round.
In general, the FL training task requires hundreds of rounds to reach target accuracy.

The details of FedAvg are shown in Algorithm~\ref{alg:fedavg}. 
Here $n_i$ denotes the training data size of client $i$, the total training data size $n=\sum n_i$. The $\mathcal{B}_i$ is the batches of training data of client $i$, $\eta$ is the learning rate of the local solver. 
There have two key hyperparameters in FedAvg, the first is the number of participating clients $K$ in each round, or the participation rate $K/N$. 
For the IID setting, a higher participation rate can improve the convergence rate, but for the non-IID setting, a small participation rate is recommended to alleviate the straggler's effect~\cite{Li2020On}. 
The second is the local epoch $E$, an appropriately large $E$ can increase the convergence speed of the global optimization and reduce the communication requirement~\cite{Li2020On}.
However, an excessively large $E$ will increase the discrepancy between the local optimization solutions and the global model, which will lead to the federated training procedure be volatile and yields suboptimal results.

\begin{algorithm}[t]
	\caption{Federated Averaging (FedAvg)}
	\label{alg:fedavg}
	\begin{algorithmic}[1]
		\Procedure {FL Server Training}{}
		\State Initialize global model $\bm{w}_0$, then $\bm{w}_1 \leftarrow \bm{w}_0$.
		\For{each communication round $t=1,2,...,T$}
			\State $S_t \leftarrow$ Server selects a random subset of $K$ clients.
			\State Server broadcasts $\bm{w}_t$ to all selected clients.
			\For{each activate client $i \in S_t$ parallelly}
				\State $\Delta\bm{w}_{t+1}^i\leftarrow$~\textbf{ClientUpdate($i$,~$\bm{w}_t$)}.
			\EndFor
			\State $\bm{w}_{t+1}\leftarrow\bm{w}_t+\sum_{i \in S_t} \frac{n_i}{n} \Delta\bm{w}_{t+1}^i$
		\EndFor
		\EndProcedure

		\Function{ClientUpdate}{$i$,~$\bm{w}$}
		\State $\hat{\bm{w}}\leftarrow\bm{w}$.
		\For{each local epoch $e=1,2,...,E$}
		\State $\bm{w}\leftarrow\bm{w}-\eta\nabla L(b;\bm{w})$ for local batch $b \in \mathcal{B}_i$.
		\EndFor
		\State \Return$\Delta\bm{w}\leftarrow\bm{w}-\hat{\bm{w}}$
		\EndFunction
	\end{algorithmic}
\end{algorithm}

\subsection{Clustered Federated Learning}\label{sec:background-CLF}
As mentioned earlier, one of the main challenges in the design of the large-scale FL system is statistical heterogeneity (e.g. non-IID, size imbalanced, class imbalanced)~\cite{li2020federated, bonawitz2019towards, sattler2019robust, briggs2020federated, duan2020self, wang2021addressing}. 
The conventional way is to train a consensus global model upon incongruent data, which yields unsatisfactory performance in the high heterogeneity setting~\cite{zhao2018federated}. 
Instead of optimizing a consensus global model, CFL divides the optimization goal into several sub-objectives and follows a \textit{Pluralistic Group} architecture~\cite{lee2020tornadoaggregate}. CFL maintains multiple groups (or clusters) models, which are more specialized than the consensus model and achieve high accuracy. 

Sattler \textit{et al.} propose the first CFL-based framework~\cite{sattler2020clustered}, which recursively separates the two groups of clients with incongruent descent directions. The authors further propose~\cite{sattler2020byzantine} to improve the robustness of CFL-based framework in the byzantine setting.
However, the recursive bi-partitioning algorithm is computationally inefficient and requires multiple communication rounds to completely separate all incongruent clients.
Furthermore, since the participating client in each round is random, this may cause the recursive bi-partitioning algorithm to fail.
To improve the efficiency of CFL, Ghosh \textit{et al.} propose IFCA~\cite{ghosh2020efficient}, which randomly generates cluster centers and divides clients into clusters that will minimize their loss values.
Although the model accuracy improvements of IFCA are significant, IFCA needs to broadcast all group models in each round and is sensitive to the initialization with probability success.
FeSEM~\cite{xie2020multi} uses a $\ell_2$ distance-based stochastic expectation maximization (EM) algorithm instead of distance-based neighborhood methods. However, $\ell_2$ distance often suffer in high-dimension, low-sample-size (HDLSS) situation which is known as distance concentration phenomenon in high dimension~\cite{sarkar2019perfect}.
The distance concentration of $\ell_p$ will cause the violation of neighborhood structure~\cite{radovanovic2010existence}, which has adverse effects on the performance of pairwise distance-based clustering algorithms such as K-Means, k-Medoids and hierarchical clustering~\cite{sarkar2019perfect}.
Similarity to the cosine similarity-based clustering method~\cite{sattler2020clustered}, Briggs \textit{et al.} propose an agglomerative hierarchical clustering method named FL+HC~\cite{briggs2020federated}.
It relies on iterative calculating the pairwise distance between all clusters, which is computationally complex. 
Note that, all the above CFL frameworks assume that all clients will participate in the clustering process and no newcomer devices, which is unpractical in a large-scale federated training system.

\section{FlexCFL}\label{sec:design}

\subsection{Motivation}
Before introducing FlexCFL, we show a toy example to illustrate the motivation of our work. 
To study the impacts of statistical heterogeneity, model accuracy, and discrepancy, we implement a convex multinomial logistic regression task train with FedAvg on MNIST~\cite{lecun1998gradient} following the instructions of~\cite{li2018federated}. 
We manipulate the statistical heterogeneity by forcing each client to have only a limited number of classes of data. 
All distributed client data is randomly sub-sampled from the original MNIST dataset without replacement, and the training set size follows a power law. 

\begin{figure*}[t]
	\centering
	\includegraphics[scale=0.9]{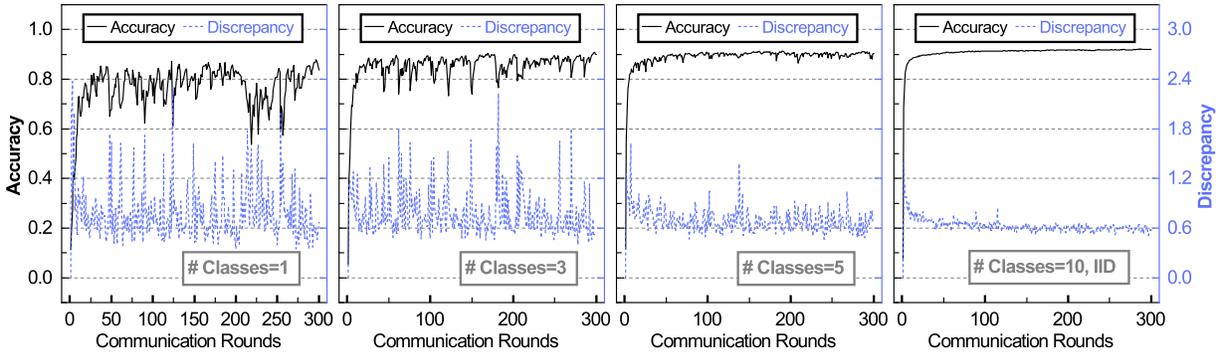}
	\caption{A FedAvg training procedure on three non-IID MNIST datasets and one IID MNIST dataset to illustrate the effects of statistical heterogeneity on model accuracy and discrepancy. 
	From left to right, the number of classes of training data per client increase, which means the degree of data heterogeneity decreases. The discrepancy is defined in Equation~\eqref{for:discrepancy}.}
	\label{fig:motivation}
\end{figure*}

We build three non-IID and one IID MNIST datasets with different class limitations.
The total client size $N=1000$, the number of clients selected in each round $K=20$, the local mini-batch size is $10$, the local epoch $E=20$, the learning rate $\eta=0.03$.
We run $T=300$ rounds of training and evaluated the global model on the local test set.
Note that although we use the same test data in each round, we did not fine-tune the model structure based on the testing results, so we ignore the possibility of test data leakage~\cite{dwork2015reusable}.

The experimental results are shown in Fig.~\ref{fig:motivation}. 
Where the left y-axis is the testing top-1 classification accuracy, and the right y-axis is the arithmetic mean of norm difference between the clients' model weights and the latest global model weights. Specifically, the discrepancy in communication round $t$ is defined as:
\begin{equation}
	\label{for:discrepancy}
	Discrepancy(t) \triangleq \frac{1}{|S_t|} \sum_{i \in S_t}\|\bm{w}_i-\bm{w}_t\|.
\end{equation}

As shown by the blue lines in Fig.~\ref{fig:motivation}, the discrepancy is relaxed as the class number limitation of each client's data increases, which means the high heterogeneity will make the trained models prone to diverge. 
Moreover, high data heterogeneity also hurt the convergence rate of training and the model accuracy, which are shown by the black lines in Fig~\ref{fig:motivation}. The accuracy curves become more fluctuant with the increase of data heterogeneity. 
The quantitative results which are shown in TABLE~\ref{tab:motivation} support our observations. 
As the heterogeneity increases, the variance of the discrepancy significantly decreases by 92.5\% (from 0.11 to 0.0082), and the max accuracy is increased +4.3\% (from 87.9\% to 92.2\%). 
The number of the required communication rounds to reach 85\% accuracy is also significantly reduced in the IID setting, which means faster convergence and less communication consumption.

Li \textit{et al.}~\cite{Li2020On} theoretically prove that the heterogeneity will also slow down the convergence rate. 
Instead of optimizing a complex global goal, why not divide it into several sub-optimization goals? 
Based on the idea of divide and conquer, many derived CFL frameworks~\cite{briggs2020federated, ghosh2020efficient, sattler2020byzantine, sattler2020clustered} manage to group the training of clients based on the proximities between their local optimizations.
While the above CFL frameworks can achieve significant performance improvements, unfortunately, there are three emerging problems. 
First, the runtime clustering mechanisms need to calculate the proximity in each round, which raises the additional computational overhead. 
Second, the existing CFL frameworks lack a mechanism to deal with newcomers.
Last, there are few discussions or solutions about the data distribution shift situation in clustered federated training.
For that reasons, our proposed FlexCFL uses a static grouping strategy combined with client migration.
We further experimentally illustrate the above problems in Section~\ref{sec:shift}.

\begin{table}[]
	\scriptsize
	\caption{Quantitative results of FedAvg training based on non-IID and IID MNIST with different \#classes/client.}
	\centering
	\label{tab:motivation}
	\begin{tabular}{cccccc}
		\hline
		\multicolumn{1}{c|}{\multirow{2}{*}{\# Classes}} & \multicolumn{2}{c|}{Discrepancy} & \multicolumn{2}{c|}{Accuracy} & \# Round to Reach \\ \cline{2-5}
		\multicolumn{1}{c|}{} & \multicolumn{1}{c|}{Mean} & \multicolumn{1}{c|}{Variance} & \multicolumn{1}{c|}{Max} & \multicolumn{1}{c|}{Median} & Target Acc-85\% \\ \hline
		1			&	0.767	&	0.11	& 87.9\% & 79.3\% & 39 \\
		3			&	0.767	&	0.073	& 90.9\% & 87.0\% & 14 \\
		5			&	0.685	&	0.021	& 91.3\% & 90.0\% & 10 \\
		10 (IID)	&	0.627	&	0.0082	& 92.2\% & 91.5\% & 4 \\ \hline
	\end{tabular}
\end{table}

\subsection{Framework Overview}
Our proposed framework is inspired by CFL~\cite{sattler2020clustered}, which clusters clients by a recursive bi-partitioning algorithm. 
CFL assumes the clients have incongruent risk functions (e.g. randomly swapping out the labels of training data), and the goal of clustering is to separate clients with incongruent risk functions.
FlexCFL borrows the client clustering idea of CFL.
But unlike previous CFL-based frameworks~\cite{sattler2020clustered, sattler2020byzantine, ghosh2020efficient, xie2020multi, liu2020client}, our grouping strategy is static, which avoids rescheduling clients for each round.
More significantly, we propose the Euclidean distance of Decomposed Cosine similarity (EDC) for efficient clustering of high dimensional direction vectors, which can be regarded as a decomposed variant of MADD~\cite{sarkar2019perfect}.

\begin{figure}[t]
	\centering
	\includegraphics[scale=0.9]{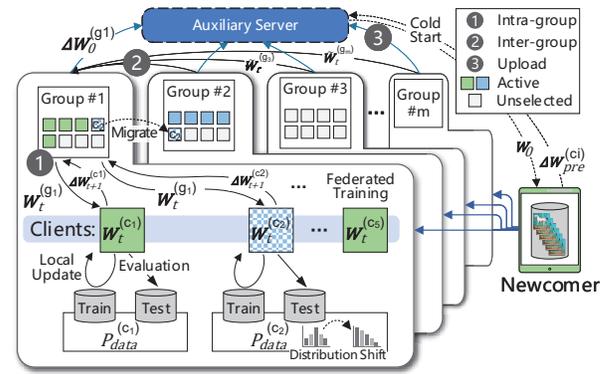}
	\caption{An overview of FlexCFL.}
	\label{fig:overview}
\end{figure}

Before we go into more detail, we first show the general overview of the FlexCFL. 
The federated training procedure of FlexCFL is shown in Fig.~\ref{fig:overview}.
Note that Fig.~\ref{fig:overview} is a schematic diagram, it does not represent the physical deployment of connections or devices.
In practice, the clients can be connected mobiles or IoT devices~\cite{abdulrahman2020fedmccs}, the auxiliary server can be deployed on cloud (act as the FL server), and the group can be deployed on cloud or the mobile edge computing (MEC) server.
To ease the discussion, we assume that all groups are deployed on the same cloud server, and all clients are connected mobile devices.

FlexCFL contains one auxiliary server, a certain amount of groups, and multiple clients, each of them maintains the latest model and the latest update for this model.
Each client and group have a one-to-one correspondence, but there are also possible to have a group without clients in a communication round (empty group) or a client is not in any groups (e.g. a newcomer joins the training later).
The newcomer device uses a cold start algorithm to determine the assigned group (we call the unassigned client is cold), we will explain the details of the cold start algorithm later.
For the dataset, each client $c_i$ has a training set and a test set according to its local data distribution $p_{data}^{(c_i)}$. 

As shown in the Fig.~\ref{fig:overview}, FlexCFL has three model transmission processes, including intra-group aggregation (\ding{192}), inter-group aggregation (\ding{193}), and optimization gradient upload (\ding{194}). 
First, the auxiliary server determines the initial model and optimization direction of each group through clustering. 
Then, each group federally trains on a certain set of clients based on their training data to optimize its group model, and evaluates the group model on the same set of clients based on its test data.
Specifically, each group broadcasts its model parameters to their clients and then aggregates the updates from these clients using \textit{FedAvg} in parallel, we call this aggregation procedure the intra-group aggregation.
After all federated trainings of groups are complete, the models are aggregated using a certain weight, which we call inter-group aggregation.
Note that, unlike the previous FedAvg-based frameworks~\cite{mcmahan2017communication, smith2017federated, bonawitz2017practical, li2018federated, duan2019astraea}, the optimization gradients in the server will not be updated or broadcast to all clients or groups in each round, we only maintain these gradients for the cold start of newcomers.

\textbf{Training:} The details of the training procedure of FlexCFL are shown in Algorithm~\ref{alg:flexcfl}, where $m$ controls the number of groups.
Before training, the server first initializes the global model and group models to the same initial weights $\bm{w}_0$. 
At the beginning of each round, a random subset of clients is selected to participate in this round of training (line 7).
Each group trains the model using FedAvg and get a temporary group model $\tilde{\bm{w}}_{t+1}^{(g_j)}$ (line 10), the group parameters are refreshed after the inter-group aggregation (line 13). 
Some details not shown in Algorithm~\ref{alg:flexcfl} is that the groups and clients need to tackle cold start issues before training. We will discuss our strategies for cold start in the next two sections. 
When $\eta_g = 0$, our framework is pluralistic~\cite{lee2020tornadoaggregate} like the previous CFL works (e.g.~\cite{ghosh2020efficient, sattler2020clustered}).
The inter-group aggregation ($\eta_g > 0$) constructs a novel \textit{semi-pluralistic} architecture for CFL-based frameworks, and we will explore it experimentally in Section~\ref{sec:evaluation}.

\begin{algorithm}[t]
	\caption{FlexCFL}
	\label{alg:flexcfl}
	\scriptsize
	\begin{flushleft}
		\textbf{Input:} Clients set $\mathcal{C} \leftarrow \{c_1, c_2, ..., c_n\}$, groups set $\mathcal{G} \leftarrow \{g_1, g_2, ..., g_m\}$, \\
		initial group parameters set $\mathcal{W}^{(G)}_0 \leftarrow \{\bm{w}^{(g_1)}_0, \bm{w}^{(g_2)}_0, ..., \bm{w}^{(g_m)}_0\}$, \\
		$g_j.clients \leftarrow \{c_i | c_i \text{~is in group~}g_j, \forall c_i \in \mathcal{C} \}$,
		initial model parameters $\bm{w}_0$, number of communication rounds $T$, number of selected clients per round $K$, inter-group learning rate $\eta_g$, proximal hyperparmeter $\mu$, distribution shift threshold $\uptau$.\\
		\textbf{Output:} Updated group model parameters $\mathcal{W}^{(G)}_T$. 
	\end{flushleft}
	\begin{algorithmic}[1]
		\Procedure {FlexCFL Training}{}
		\State $\mathcal{W}^{(G)}_1 \leftarrow \{\bm{w}^{(g_1)}_1, \bm{w}^{(g_2)}_1, ..., \bm{w}^{(g_m)}_1\} \leftarrow$ initialized $\mathcal{W}^{(G)}_0$ by $\bm{w}_0$.
		\For{each communication round $t=1,2,...,T$}
		\For{each client $c_i$ has been cold start parallelly}
		\If{Wasserstein distance of the change of $P_{data}^{(c_i)} > \uptau_i$}
		\State Redo the cold start of $c_i$ locally.~//Client migration.
		\EndIf
		\EndFor
		\State $S_t \leftarrow$ Server selects a random subset of $K$ clients.
		\For{each group $g_j$ in $\mathcal{G}$ parallelly}
		\State $S_t^{(g_j)} \leftarrow \{c_i|c_i \in g_j.clients, \forall c_i \in S_t\}$.
		\State $\tilde{\bm{w}}_{t+1}^{(g_j)} \leftarrow$~\textbf{IntraGroupUpdate($S_t^{(g_j)}$, $\bm{w}_{t}^{(g_j)}$)}.
		\State $\tilde{\mathcal{W}}_{t+1}^{(G)} \leftarrow \{\tilde{\bm{w}}_{t+1}^{(g_1)}, \tilde{\bm{w}}_{t+1}^{(g_2)},...,\tilde{\bm{w}}_{t+1}^{(g_m)}\}$.
		\EndFor
		\For{each group $g_j$ in $\mathcal{G}$ parallelly}
		\State $\mathcal{W}_{t+1}^{(G)} \leftarrow$~\textbf{InterGroupAggregation($\tilde{\mathcal{W}}_{t+1}^{(G)}$, $\eta_g$)}.
		\EndFor
		\EndFor
		\EndProcedure
		
		\Function{IntraGroupUpdate}{$S_t, \bm{w}_t$}
		\If{$S_t$ is $\emptyset$} \Return $\bm{w}_t$.~//Empty group.
		\EndIf
		\State $\bm{w}_{t+1} \leftarrow FedAvg(S_t, \bm{w}_t)$.~//Ref.~Algorithm~\ref{alg:fedavg}.
		\State \Return $\bm{w}_{t+1}$
		\EndFunction
		
		\Function{InterGroupAggregation}{$\tilde{\mathcal{W}}_{t+1}, \eta_g$}
		\For{each group parameter $\tilde{\bm{w}}_{t+1}^{(g_j)}$ in $\tilde{\mathcal{W}}_{t+1}$ parallelly}
		\State  $\Delta\tilde{\bm{w}}_{t+1}^{(g_j)} \leftarrow \eta_g 
		\sum_{l\neq j} \frac{\tilde{\bm{w}}_{t+1}^{(g_l)}}{\|\tilde{\bm{w}}_{t+1}^{(g_l)}\|}$, then $\bm{w}_{t+1}^{(g_j)} \leftarrow \tilde{\bm{w}}_{t+1}^{(g_j)} + \Delta\tilde{\bm{w}}_{t+1}^{(g_j)}$.
		\EndFor
		\State \Return $\mathcal{W}_{t+1} \leftarrow \{\bm{w}^{(g_1)}_{t+1}, \bm{w}^{(g_2)}_{t+1}, ..., \bm{w}^{(g_m)}_{t+1}\}$
		\EndFunction
	\end{algorithmic}
\end{algorithm}

\subsection{Group Cold Start}

Based on the guideline of CFL, we manage to group the trainings of clients based on the proximities between their local optimizations.
But there are two following questions: (1) how to measure the proximity or distance, and (2) how to determine the optimization goals of each group before training.
For the choice of measure, a heuristic way is to use the $\ell_2$ distance between models, which has miserable performance as we mentioned in Section~\ref{sec:background-CLF}.
The loss value can be used as a surrogate for the proximity of difference domains~\cite{mohri2019agnostic}, but the huge computational overhead involved in model inferencing.
The cosines similarity between the gradients calculated by backpropagation or the updates of model parameters is an alternative measure.
The cosine similarity between the updates of any two clients $c_i$ and $c_j$ is defined by:
\begin{equation}
	\mathcal{S}(i,j) \triangleq \frac{<\Delta\bm{w}_t^{(c_i)}, \Delta\bm{w}_t^{(c_j)}>}{\|\Delta\bm{w}_t^{(c_i)}\|~\|\Delta\bm{w}_t^{(c_j)}\|},
\end{equation}
and the pairwise cosine similarity matrix $\mathcal{M} \subset \mathbb{R}^{n \times n}$ can be written with cosine similarity kernel $K$ as follows:

\begin{equation}\label{eq:cossim}
	\mathcal{M} = K(\Delta \bm{W},\Delta \bm{W}),~\mathcal{M}_{ij} = \mathcal{S}(i,j).
\end{equation}

The computational complexity of calculating $\mathcal{M}$ is $O\big(n^2d^2_{\bm{w}}\big)$, $n$ and $d_{\bm{w}}$ are the number of clients and number of parameters, respectively.
We assume the all parameters updates $\Delta\bm{w}_t$ are flattened row vectors, so $\Delta\bm{w}_t \subset \mathbb{R}^{1 \times d_{\bm{w}}}$.
In general, $d_{\bm{w}}$ is huge and $d_{\bm{w}} \gg n$ (HDLSS), which make the pairwise cosine similarity-based clustering methods computationally inefficient.
Unlike $\ell_p$ distance, the expectation of $\mathcal{S}$ asymptotically remains constant as dimensionality increases~\cite{radovanovic2010existence}, which is friendly to the clustering in high dimensional data.
Unfortunately, $\mathcal{S}$ is not suitable for low-dimensional situations because the variance of it is $O(1/d_{\bm{w}})$.
Therefore, we extend the data-driven method MADD~\cite{sarkar2019perfect} to our cosine similarity-based case, so we can reduce the observation bias by using the mean of residuals of $\mathcal{S}$, fox example:

\begin{equation}\label{eq:MADC}
	MADC(i,j)=\frac{1}{n-2} \sum_{z \neq i,j} |\mathcal{S}(i, z) - \mathcal{S}(j, z)|.
\end{equation}

The above dissimilarity measure is based on the Mean of Absolute Differences of pairwise Cosine similarity, so we call it MADC.
However, MADC and $\mathcal{M}$ have the same computation complexity and both are proximity measures, which means they cannot be applied for some efficient Euclidean distance-based clustering algorithms. These motivate us to develop a variant of MADC called Euclidean distance of Decomposed Cosine similarity (EDC), which is defined by:

\begin{equation}\label{eq:EDC}
	\begin{aligned}
		EDC(i,j)=&~\frac{1}{m} \sqrt{\sum_{v \in \bm{V}} {(\mathcal{S}(i, v) - \mathcal{S}(j, v))}^2}, ~or\\
		EDC(i,j)=&~\frac{1}{m} \| K(\Delta\bm{w}_t^{(c_i)}, \bm{V}^T) - K(\Delta\bm{w}_t^{(c_j)}, \bm{V}^T)\|, \\
		\bm{V}=&~SVD(\Delta \bm{W}^T, m), \bm{V} \subset \mathbb{R}^{d_{\bm{w}} \times m}.
	\end{aligned}
\end{equation}

Instead of calculating the pairwise similarity, EDC first decomposes the updates of models into $m$ directions by using truncated Singular Value Decomposition (SVD) algorithm~\cite{golub1971singular}, so then only the similarities between the updates and these directions will be calculated. It is worth noting that the complexity of truncated SVD is only $O(2m^2d_{\bm{w}})$ for $d_{\bm{w}} \gg m$ and hence the computational complexity of EDC is $O(m^2d_{\bm{w}}^2)$. Some previous works (i.e.~\cite{sattler2020clustered, briggs2020federated}) calculate the pairwise cosine similarity based on all participants and $n$ is usually hundreds or thousands, so $O(m^2d_{\bm{w}}^2) \ll O(n^2d_{\bm{w}}^2)$. Furthermore, \cite{sattler2020clustered, briggs2020federated} leverage the hierarchical clustering strategies, which are recursively and computationally expensive.

To determine the optimization goals of each group, FlexCFL clusters the parameter updates directions of clients into $m$ groups using K-Means++~\cite{arthur2006k} algorithm based on EDC.
The main advantage of our clustering approach is that it is unsupervised, and we can divide the global optimization function into $m$ sub-optimization functions regardless of whether there have incongruent optimization goals.
In other words, FlexCFL performs a low-dimensional embedding of the local updates matrix $\Delta \bm{W}$, following by K-Means++ clustering.
Our calculation of client clustering is following the calculation of the similarity matrix, which means the above calculations only require one round of communication.
We call the combination of the above two processes as group cold start, and the details are shown in Algorithm~\ref{alg:groupcoldstart}. For comparison, we also provide the MADC version of FlexCFL, which is clustered using the hierarchical strategy with the complete linkage.

After the server performs the group cold start, the optimization direction of group $j$, which is measured by $\Delta \bm{w}_0^{(g_j)}$ is determined, and the $\alpha m$ clients participating in this process are assigned. 
We leverage the centric means of groups to measure the clustering validity index like within-cluster sum-of-squares criterion.
The optimization gradients will be uploaded to the auxiliary server for cold start of the newcomers.
Another improvement of our algorithm is that we only select a subset of clients to participate in the pre-training and decomposition process.
Because training all clients to cluster their updated directions is not communication-friendly and practically achievable, we control the scale of pre-training by hyperparameter $\alpha$ and the number of pre-training clients is set to $\alpha m$.
Our implementation is more practical because it is difficult to satisfy that all clients are active until they complete the training in the large-scale FL systems. For example, the drop-out may occur due to the network jitter.

\begin{algorithm}[t]
	\caption{Group Cold Start}\label{alg:groupcoldstart}
	\scriptsize
	\begin{flushleft}
		\textbf{Input:} Clients set $\mathcal{C}$, number of group $m$, global initial model $\bm{w}_0$, pre-training scale hyperparameter $\alpha$. \\
		\textbf{Output:} Groups set $\mathcal{G}$, set of group parameters $\mathcal{W}_0^{(G)}$, set of group updates $\Delta \mathcal{W}_0^{(G)}$.
	\end{flushleft}
	\begin{algorithmic}[1]
		\Procedure {Group Cold Start}{}
		\State $S \leftarrow$ Server selects a random subset of $\alpha*m$ clients.
		\State Server broadcasts $\bm{w}_0$ to all selected clients.
		\For{each client $c_i$ in $S$ parallelly}
		\State $\Delta \bm{w}_0^{(c_i)} \leftarrow$~$flatten($ \textbf{ClientUpdate($i, \bm{w}$)} $)$.~//Ref.~Algorithm~\ref{alg:fedavg}.
		\EndFor
		\State $\Delta \bm{W} \leftarrow [\Delta \bm{w}_0^{(c_1)}, \Delta \bm{w}_0^{(c_2)}, \dots, \Delta \bm{w}_0^{(c_{\alpha*m})}]$.
		
		\If{MADC}:
		\State $\mathcal{M} \leftarrow K(\Delta \bm{W},\Delta \bm{W})$.~//Ref.~Eq.~(\ref{eq:cossim})
		\State Proximity matrix $\mathcal{M}_p \leftarrow$ \textbf{Calculate MADC($\mathcal{M}$)}.~//Ref.~Eq.~(\ref{eq:MADC})
		\State $[g_1.clients, \dots, g_m.clients] \leftarrow$ Hierarchical Clustering($\mathcal{M}_p, m$).
		\EndIf
		\If{EMD}:
		\State $V \leftarrow truncated~SVD(\Delta \bm{W}^T,m)$.
		\State Distance matrix $\mathcal{M}_d \leftarrow K(\Delta \bm{W}, V^T)$.~//Ref. ~Eq.~(\ref{eq:EDC})
		\State $[g_1.clients, \dots, g_m.clients] \leftarrow$ K-Means++($\mathcal{M}_d,m$).
		\EndIf
		
		\For{$g_j$ in $\mathcal{G} \leftarrow [g_1, \dots, g_m]$}
		\State $\bm{w}_0^{(g_j)} \leftarrow Average([\Delta \bm{w}_0^{(c_i)},\forall c_i \in g_j.clients])$.
		\State $\Delta \bm{w}_0^{(g_j)} \leftarrow \bm{w}_0^{(g_j)} - \bm{w}_0$.
		\EndFor
		\State $\mathcal{W}_0^{(G)} \leftarrow [\bm{w}_0^{(g_1)}, \dots, \bm{w}_0^{(g_m)}]$, and $\Delta \mathcal{W}_0^{(G)} \leftarrow [\Delta \bm{w}_0^{(g_1)}, \dots, \Delta \bm{w}_0^{(g_m)}]$.
		\State Server broadcasts $\mathcal{W}_0^{(G)}$ and $\bm{w}_0$ to $S$ for client migration.
		\State \Return $\mathcal{G}, \mathcal{W}_0^{(G)}, \Delta \mathcal{W}_0^{(G)}$
		\EndProcedure
	\end{algorithmic}
\end{algorithm}

\subsection{Client Cold Start and Migration}
As described before, the group cold start algorithm selects a random subset ($\alpha m $) of the clients for pre-training, so the remaining clients ($n-\alpha m$) are cold clients and are not in any groups.
Since the federated training network is dynamic, the new devices can join the training at any time, so we need to classify newcomers according to the similarity between their optimization goals and groups'.
Our client cold start strategy is to assign clients to the groups that are most closely related to their optimization direction, as shown below:
\begin{equation}
	\begin{aligned}
		\label{for:clientcoldstart}
		&g^* = \operatorname*{argmin}_j \frac{-\cos(\sphericalangle(\Delta \bm{w}_0^{(g_j)}, \Delta \bm{w}_{pre}^{(i)}))+1}{2}. \\
	\end{aligned}
\end{equation}
Suppose the newcomer $i$ joins the training network in round t, then the $\Delta \bm{w}_{pre}^{(i)}$ is the pre-training gradient of the newcomer base on the global initial model $\bm{w}_0$.
We schedule the newcomer $i$ to group $g^*$ to minimize the normalized cosine dissimilarity.
The client will store $\mathcal{W}_0^{(G)}$ and $\bm{w}_0$ locally for future client migration.
With this mechanism, FlexCFL does not need to broadcast all groups' models every round to dynamically schedule the clients, which significantly reduces communication consumption.

Although our experiments (Section~\ref{sec:effects}) prove that the static client scheduling strategy of FlexCFL is high-efficiency, this strategy cannot handle the distribution shift challenge (Section~\ref{sec:shift}).
The data distribution shift~\cite{miller2020effect} is a natural situation especially when the training devices are IoT nodes such as industrial sensors, wearables, cameras, etc. 
In order to maintain the advantages of our static clustering method, we propose a flexible migration strategy.
Before each round of training, we leverage Wasserstein distance to detection the distribution shift of the training data of all clients (except cold clients), if the Wasserstein distance exceeds the threshold $\uptau_i$, then a cold start will be scheduled (Algorithm~\ref{alg:flexcfl} line 6). The distribution shift threshold $\uptau_i$ of client $c_i$ is defined by:
\begin{equation}
	\uptau_i = \frac{0.2}{Label Size}n_i,
\end{equation}
which means that 20\% training data change will be considered as distribution shift.

In summary, the key features of our FlexCFL framework are as follows:
\begin{itemize}
	\item FlexCFL reduces the discrepancy between the joint optimization objective and sub-optimization objectives, which is unsupervised and can disengage from the incongruent risk functions assumption.
	\item The proposed framework determines the optimization objectives of groups by an efficient clustering approach based on a decomposed data-driven measure.
	\item The client scheduling mechanism of FlexCFL considers the joining of newcomer devices and the distribution shift of local data.
	
\end{itemize}

Compared with vanilla FL, FlexCFL requires additional computing resources to pre-training and communication to transmit optimization gradients of groups.
It is worth noting that these gradients only need to be transmitted once for each client.
In addition, we emphasize that the pre-training procedure does not occupy a whole communication round, the client can continue to train $E-1$ epochs and upload the parameters updates for the intra-group aggregation.

\section{Convergence Analysis}\label{sec:convergence}

We analyze convergence for FlexCFL in this section. First, we analyze the convergence of our proposed framework without inter-group aggregation (e.g. $\eta_g=0$ in Algorithm~\ref{alg:flexcfl} line 13), and then extend it to the case with inter-group aggregation.

In FlexCFL, the membership for each group is static during each training round, so we can assume that any client $k$ is allocated to group $g$.
We make the following assumptions on the local loss $F_{k,g}(\cdot)$ for any client $k$.

\begin{assumption}\label{assu1}
	For any client $k$, $F_{k,g}(\cdot)$ is convex.
\end{assumption}

\begin{assumption}\label{assu2}
	$F_{k,g}$ is $M$-Lipschitz continuous: for all $\bm{w}$ and $\bm{v}$, $\|F_{k,g}(\bm{w})-F_{k,g}(\bm{v})\| \leq M\|\bm{w}-\bm{v}\|$.
\end{assumption}

\begin{assumption}\label{assu3}
	$F_{k,g}$ is $L$-Lipschitz smooth: for all $\bm{w}$ and $\bm{v}$, $\|\nabla F_{k,g}(\bm{w})-\nabla F_{k,g}(\bm{v})\| \leq L\|\bm{w}-\bm{v}\|$.
\end{assumption}

The above assumptions have been made by many relevant works~\cite{lee2020tornadoaggregate, liu2020client, wang2019adaptive}.

\begin{definition}[Group Loss Function]\label{def1}
For any group $g\in \mathcal{G}$, the group loss function is $F_g(\cdot) \triangleq \sum_{k}p_kF_{k,g}(\cdot)$, and $\sum_{k}p_k=1$.
\end{definition}

\begin{lemma}
	Under Assumptions~\ref{assu1} to \ref{assu3}, the group loss function $F_g$ are convex, $M$-Lipschitz continuous, $L$-Lipschitz smooth for any $g$.
\end{lemma}

\textit{Proof.} This simply follows by Definition~\ref{def1}, given $F_g(\cdot)$ is a linear combination of the local loss function $F_{f,g}(\cdot)$.

Let $w_t^g$ be the model parameter maintained in the group $g$ and at the $t$-th step. Let $e$ be the current local epoch number, $e \in [0, E]$.  We assume that the $0$-th local epoch is the synchronization step, so $e$ will be reset before the start of the communication round $t$. Then the update of FlexCFL without inter-group aggregation can be described as:
	
\begin{equation}
\label{eq:fedgroup without agg}
\begin{aligned}
	\bm{w}_{t,e}^{k,g}&=
	\begin{cases}
		\bm{w}_t^g, & e=0\\
		\bm{w}_{t,e-1}^{k,g}-\eta\nabla F_{k,g}(\bm{w}_{t,e-1}^{k,g}), &e\in[1,E]
	\end{cases}\\
	\bm{w}_t^g &\triangleq \sum_{k}p_k\bm{w}_{t,E}^{k,g}
\end{aligned}
\end{equation}

Here we introduce an additional notation \textit{virtual group model} $\bm{v}_{t,e}^g$ to measure the divergence between federated training and SGD-based centralized training, which is motivated by~\cite{stich2018local, Li2020On, lee2020tornadoaggregate}. We assume that there is a virtual group model $\bm{v}_{t,e}^g$ that is centralized trained on the collection of members' data and is synchronized with the federated model in each communication round. We introduce this notion formally below.

\begin{equation}
	\bm{v}_{t,e}^{g}=
	\begin{cases}
		\bm{w}_t^g, & e=0\\
		\bm{v}_{t,e-1}^{g}-\eta\nabla F_{g}(\bm{v}_{t,e-1}^{g}), &e\in[1,E]
	\end{cases}
\end{equation}

\begin{definition}[Intra-Group Gradient Divergence]
\label{def2}
Given a certain group membership, for any $g$ and $k$, $\delta_{k,g}$ represents the gradient difference between the loss functions of client $k$ and group $p$, as expressed below:
	\begin{equation}
		\delta_{k,g} \triangleq \max_{\bm{w}}\|\nabla F_{k,g}(\bm{w})-\nabla F_g(\bm{w})\|
	\end{equation}
And the intra-group gradient divergence is defined as Eq.~(\ref{eq:intra-group gradient divergence}),
	\begin{equation}
		\label{eq:intra-group gradient divergence}
		\delta \triangleq \sum_{g\in \mathcal{G}}~\sum_{k\in g.clients} p_g p_k \delta_{k,g}
	\end{equation}
\end{definition}

\begin{lemma}[Upper bound of the divergence of $\bm{w}_{t,e}^{k,g}$]
	\label{lem2}
	Let Assumptions~\ref{assu1} to \ref{assu3} hold, the upper bound of divergence between the FlexCFL model and the virtual group model for any $t$, $e$ is given by
\begin{equation}
	\|\bm{w}_{t,e}^{k,g}-\bm{v}_{t,e}^g\| \leq \frac{\delta_{k,g}}{L}((\eta L+1)^{e}-1)
\end{equation}
\end{lemma}

\textit{Proof.} By the smoothness of $F_{k,g}(\cdot)$ and the Definition~\ref{def2}, we have

\begin{equation}
	\label{eq:bouding proof}
	\begin{aligned}
		&\|\bm{w}_{t,e}^{k,g}-\bm{v}_{t,e}^g\| \\
		&= \|\bm{w}_{t,e-1}^{k,g}-\eta\nabla F_{k,g}(\bm{w}_{t,e-1}^{k,g})- \bm{v}_{t,e-1}^g + \eta\nabla F_{g}(\bm{v}_{t,e-1}^{g})\| \\
		&\leq\| \bm{w}_{t,e-1}^{k,g} - \bm{v}_{t,e-1}^g \| + \eta \| \nabla F_{k,g}(\bm{w}_{t,e-1}^{k,g}) - \nabla F_{g}(\bm{v}_{t,e-1}^{g}) \| \\
		&\leq\| \bm{w}_{t,e-1}^{k,g} - \bm{v}_{t,e-1}^g \| + \eta \| \nabla F_{k,g}(\bm{w}_{t,e-1}^{k,g}) - \nabla F_{k,g}(\bm{v}_{t,e-1}^{g}) \| \\
		&~~~ +\eta \| \nabla F_{k,g}(\bm{v}_{t,e-1}^{g}) - \nabla F_{g}(\bm{v}_{t,e-1}^{g}) \| \\
		&\leq (\eta L +1) \| \bm{w}_{t,e-1}^{k,g} - \bm{v}_{t,e-1}^g \| + \eta \delta_{k,g}
	\end{aligned}
\end{equation}

Let $h(e)= \|\bm{w}_{t,e}^{k,g}-\bm{v}_{t,e}^g\| $, then we can rewrite Eq.~(\ref{eq:bouding proof}) as
\begin{equation}
	\begin{aligned}
		&h(e) \leq (\eta L+1)h(e-1)+\eta\delta_{k,i} \\
		&\frac{h(e)+\delta_{k,g}/L}{h(e-1)+\delta_{k,g}/L} \leq \eta L +1
	\end{aligned}
\end{equation}

Given that $h(0) = \|\bm{w}_{t,0}^{k,g}-\bm{v}_{t,0}^g\| = 0$, by induction, we have
\begin{equation}
	g(e) + \frac{\delta_{k,g}}{L} \leq \frac{\delta_{k,g}}{L}(\eta L +1)^e
\end{equation}
Therefore, Lemma~\ref{lem2} is proved. \hfill$\square$

Combing Lemma~\ref{lem2} and Eq.~(\ref{eq:fedgroup without agg}) and using Jensen's inequality we get
\begin{equation}
	\label{eq:different between group and virtual}
	\begin{aligned}
	\| \bm{w}_t^g -\bm{v}_{t,e}^g \| &\leq \sum_{k} p_k \| \bm{w}_{t,E}^{k,g} - \bm{v}_{t,e}^g \| \\
	&\leq \frac{\delta}{L}((\eta L+1)^{E}-1)
	\end{aligned}
\end{equation}

Consider the continuous of $F_g(\cdot)$ we have

\begin{equation}
	\label{eq:bound fedgroup without agg}
	\| F_g(\bm{w}_t^g) - F_g(\bm{v}_{t,e}^g) \| \leq \frac{\delta M}{L}((\eta L+1)^{E}-1)
\end{equation}

\begin{theorem}[Convergence Bound of FlexCFL without inter-group aggregation]
\label{theo1}
Let Assumption~\ref{assu1} to \ref{assu3} hold and $g,t,E,\bm{w}_t^g,\bm{v}_{t,e}^g$ be defined therein. Then the convergence bound between the federated group model and the virtual group model is $\frac{\delta M}{L}((\eta L+1)^{E}-1)$.
\end{theorem}

Then we extent Theorem~\ref{theo1} to the case where $\eta_g>0$. First we introduce $\tilde{\bm{w}}_t^g$ to represent the model parameter of group $g$ after inter-group aggregation. Then the update of FlexCFL with inter-group aggregation can be described as Eq.~(\ref{eq:fedgroup without agg}) and
\begin{equation}
	\tilde{\bm{w}}_t^g = \bm{w}_t^g+\eta_g \sum_{l\in \mathcal{G}, l \neq g}\frac{\bm{w}_t^l}{\| \bm{w}_t^l \|}
\end{equation}

We replace $\bm{w}_t^g$ in Eq.~(\ref{eq:different between group and virtual}) with $\tilde{\bm{w}}_t^g$ and derive

\begin{equation}
\begin{aligned}
	\| \tilde{\bm{w}}_t^g -\bm{v}_{t,e}^g \| &= \| \bm{w}_{t}^{g} - \bm{v}_{t,e}^g + \eta_g \sum_{l\in \mathcal{G}, l \neq g}\frac{\bm{w}_t^l}{\| \bm{w}_t^l\|} \| \\
	&\leq \| \bm{w}_{t}^{g} - \bm{v}_{t,e}^g \| + \eta_g \sum_{l\in \mathcal{G}, l \neq g} \| \frac{\bm{w}_t^l}{\| \bm{w}_t^l\|} \| \\
	&\leq \frac{\delta}{L}((\eta L+1)^{E}-1) + \eta_g (|\mathcal{G}|-1)
\end{aligned}
\end{equation}

Then by the M-Lipschitz continuous of $F_g(\cdot)$ we get the convergence bound of FlexCFL with inter-group aggregation:
\begin{equation}
	\label{eq:bound fedgroup with agg}
	\| F_g(\tilde{\bm{w}}_t^g) - F_g(\bm{v}_{t,e}^g) \| \leq \frac{\delta M}{L}((\eta L+1)^{E}-1) + \eta_g (|\mathcal{G}|-1)
\end{equation}

Note that, Eq.~(\ref{eq:bound fedgroup with agg}) degrades to Eq.~(\ref{eq:bound fedgroup without agg}) when $\eta_g=0$ or $|\mathcal{G}|=1$, which means the learning rate of inter-group aggregation is 0 (disabled) or there is only one group in FlexCFL (be degraded to FedAvg framework).

\section{Evaluation}\label{sec:evaluation}
In this section, we present the experimental results for FlexCFL framework.
We show the performance improvements of our framework on four open datasets.
Then we demonstrate the effectiveness of FlexCFL, which includes the clustering algorithm (group cold start), the newcomer assignment algorithm (client cold start), client migration strategy.
Our implementation is based on Tensorflow~\cite{abadi2016tensorflow}, and all code and data are publicly available at \textit{https://github.com/morningD/FlexCFL}.
To ensure reproducibility, we fix the random seeds of the clients' selection and initialization.

\subsection{Experimental Setup}
We evaluate FlexCFL on four federated datasets, which including three image classification tasks and a synthetic dataset. 
In this section, we adopt the same notation for federated learning settings as Section~\ref{sec:design} and as~\cite{li2018federated}: the local epoch $E=10$, the number of selected clients per round $K=20$, the pre-training scale $\alpha=20$. The local solver is a mini-batch SGD with $B=10$. Besides, the learning rate $\eta$ and FedProx hyperparameter $\mu$ in our experiments are consistent with the recommended settings of~\cite{li2018federated}. 

\noindent\textbf{Datasets and Models}
\begin{itemize}
\item MNIST~\cite{lecun1998gradient}. A 10-class handwritten digits image classification task, which is divided into 1,000 clients, each with only two classes of digits. We train a convex multinomial logistic regression (MCLR) model and a non-convex multilayer perceptron (MLP) model based on it. The MLP has one hidden layer with 128 hidden units.
\item Federated Extended MNIST (FEMNIST)~\cite{cohen2017emnist}. A handwritten digits and characters image classification task, which is built by resampling the EMNIST~\cite{cohen2017emnist} according to the writer and downsampling to 10 classes ('a'-'j'). We train a MCLR model, a MLP model (one hidden layer with 512 hidden units), a CNN model (6 convolution layers) based on it.
\item Synthetic. It's a synthetic federated dataset proposed by \textit{Shamir et. al}~\cite{shamir2014communication}. Our hyperparameter settings of this data-generated algorithm are $\alpha=1, \beta=1$, which control the statistical heterogeneity among clients. We study a MCLR model based on it.
\item FashionMNIST~\cite{xiao2017fashion}. A 28*28 grayscale images classification task, which comprises 70,000 fashion products from 10 categories. The data partitions on each client refer to~\cite{li2020ditto}. We study a CNN model with 2 convolution layers based on it.
\end{itemize} 
The statistics of our experimental datasets and models are summarized in TABLE~\ref{tab:dataset}.

\begin{table}[h]
	\scriptsize                                                    
	\caption{Statistics of Federated Datasets and Models.}                 
	\centering
	\label{tab:dataset}
	\begin{tabular}{lllll}
		\hline
		Dataset & Devices & Samples & Model & $d_{\bm{w}}$ \\ \hline
		\multirow{3}{*}{MNIST} & \multirow{3}{*}{1,000} & \multirow{3}{*}{69,035} & MCLR & 7,850 \\ \cline{4-5} 
		&                   &                   & MLP & 101,770 \\ \cline{4-5} 
		&                   &                   & CNN & 50,186 \\ \hline
		\multirow{3}{*}{FEMNIST} & \multirow{3}{*}{200} & \multirow{3}{*}{18,345} & MCLR & 7,850 \\ \cline{4-5} 
		&                   &                   & MLP & 407,050 \\ \cline{4-5} 
		&                   &                   & CNN & 325,578 \\ \hline
		Synthetic(1,1)& 100 & 75,349 & MCLR & 610 \\ \hline
		FashionMNIST & 500 & 72,505 & CNN & 3,274,634 \\ \hline
	\end{tabular}
\end{table}

\noindent\textbf{Baselines}
\begin{itemize}
\item FedAvg~\cite{mcmahan2017communication}: the vanilla FL framework.

\item FedProx~\cite{li2018federated}: A popular federated learning optimizer which adds a quadratic penalty term to the local objective.

\item IFCA~\cite{ghosh2020efficient}: An CFL framework that minimizes the loss functions while estimating the cluster identities.

\item FeSEM~\cite{xie2020multi}: An $\ell_2$ distance-based CFL framework that minimizes the expectation of discrepancies between clients and groups stochastically.

\item FedGroup~\cite{duan2020fedgroup}: A preliminary version of FlexCFL based on EDC without client migration. The comparisons between EDC and MADC have been given in our previous work~\cite{duan2020fedgroup}, these content won't be covered here.
\end{itemize} 

Although our experiments show that using FedAvg as the intra-group aggregation strategy of IFCA and FeSEM can achieve better results, we still use simply averaging consistently with the original description.

\begin{table*}[t]
	\small
	\centering
	\caption{Comparisons with FedAvg\cite{mcmahan2017communication}, FedProx\cite{li2018federated}, IFCA\cite{ghosh2020efficient}, FeSEM\cite{xie2020multi}, FlexCFL, FlexCFL with $\eta_g=0.1$ on MNIST, FEMNIST, Synthetic, FashionMNIST without distribution shift. Ablation studies of FlexCFL: Random Cluster Centers (RCC), Randomly Assign Cold (i.e. newcomers) Clients (RAC). The accuracy improvements $\uparrow$ are calculated relative to the $\ell_2$ distance-based CFL framework FeSEM. Local Epoch $E=10$.}
	\begin{tabular}{|l|cc|cc|cccc|}
		\hline
		Dataset-Model & FedAvg & FedProx & IFCA & FeSEM & RCC & RAC & FlexCFL & FlexCFL-$\eta_g$ \\ \hline
		MNIST-MCLR & $89.4$ & $90.9$ & $94.2$ & $84.6$ & $93.4$ & $88.2$ & $\bm{95.8}(\uparrow11.2)$ & $95.2$ \\
		MNIST-MLP & $92.7$ & $94.5$ & $95.8$ & $89.8$ & $96.2$ & $91.6$ & $\bm{97.0}(\uparrow7.2)$ & $96.6$ \\
		MNIST-CNN & $97.3$ & $94.6$ & $97.8$ & $92.0$ & $98.3$ & $95.3$ & $\bm{99.0}(\uparrow7.0)$ & $98.8$ \\
		FEMNIST-MCLR & $74.9$ & $76.7$ & $85.8$ & $43.7$ & $88.8$ & $67.0$ & $\bm{90.1}(\uparrow46.4)$ & $86.0$ \\
		FEMNIST-MLP & $79.6$ & $79.4$ & $87.9$ & $57.9$ & $\bm{93.9}$ & $69.4$ & $92.9(\uparrow35.0)$ & $88.9$ \\
		FEMNIST-CNN & $95.2$ & $96.3$ & $98.1$ & $57.3$ & $98.7$ & $86.1$ & $\bm{98.7}(\uparrow41.4)$ & $98.5$ \\
		Synthetic(1,1)-MCLR & $66.9$ & $80.7$ & $\bm{91.3}$ & $77.0$ & $86.0$ & --- & $85.6(\uparrow8.6)$ &  $80.3$ \\
		FashionMNIST-CNN & $89.2$ & $89.1$ & $90.9$ & $81.3$ & $91.2$ & $85.0$ & $\bm{92.6}(\uparrow11.3)$ & $92.4$ \\ \hline
	\end{tabular}
	\label{tab:result1}
\end{table*}

\noindent\textbf{Evaluation Metrics} 

Since each client has a local test set in our experimental setting, we evaluate its corresponding group model based on these data. 
For example, in FedAvg and FedProx, we evaluate the global model based on the test set of all clients. 
And in FlexCFL and IFCA, we evaluate the group model based on the test set of the clients in this group.
We use top-1 classification accuracy to measure the performance of the classifiers.
Given CFL-based frameworks have multiple accuracies of groups with different sizes, we use a "weighted" accuracy to measure the overall performance and the weight is proportional to the test data size of each group.
In fact, the "weighted" accuracy is equivalent to the sum of the misclassified sample count in all groups divided by the total test size.

To make our results more comparable, the test clients of the group model are all the clients historically assigned to this group.
Also, we discard the test accuracy until all clients are included in the test.
Note that the heterogeneity will affect the convergence, resulting in greater fluctuations in model accuracy during the training process.
Therefore, we assume the early stopping~\cite{yao2007early} strategy is applied and we regard the maximum test accuracy during training as the final score. The number of groups of all CFL-based frameworks remains the same for each dataset. 

\begin{figure*}[t]
	\vspace{-8mm}
	\centering
	\subfloat[MNIST-MCLR]{
		\hspace{-8mm}
		\includegraphics{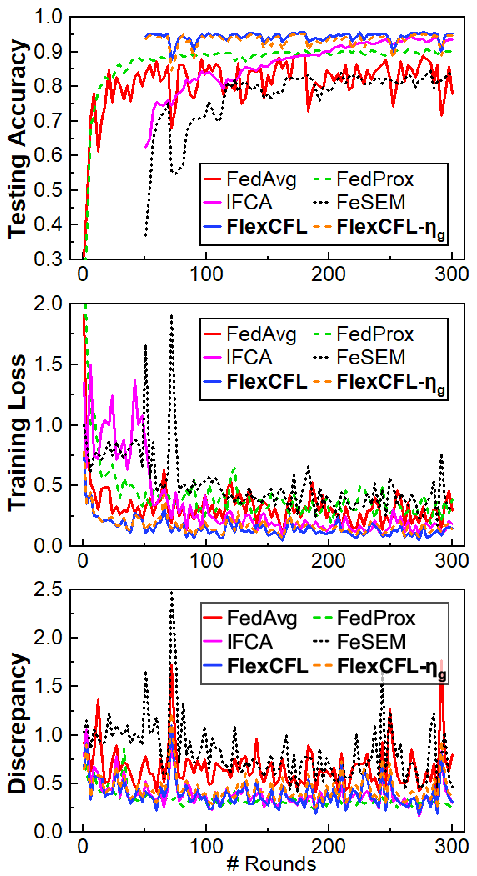}
		\label{fig:mnist1}
		\hspace{-8mm}
	}
	\subfloat[MNIST-MLP]{
		\includegraphics{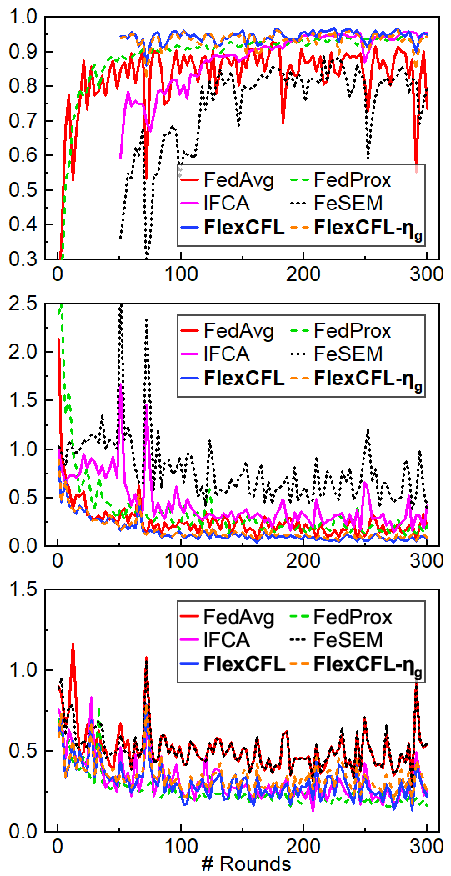}
		\label{fig:mnist2}
		\hspace{-8mm}
	}
	\subfloat[MNIST-CNN]{
		\includegraphics{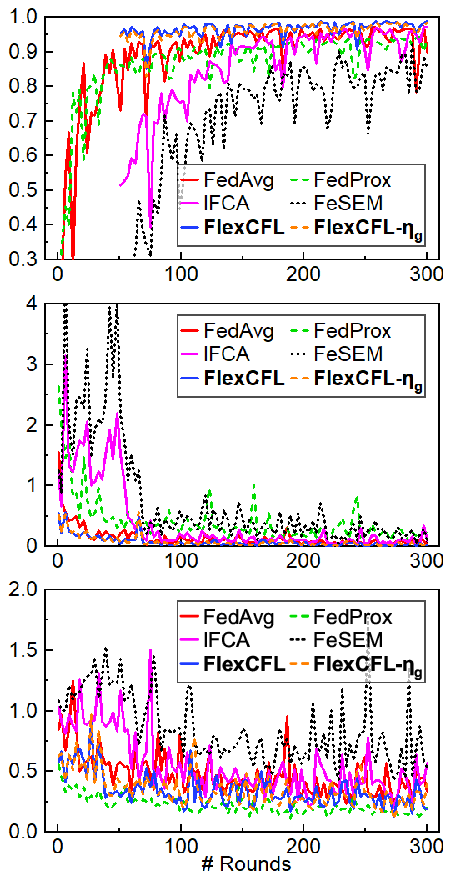}
		\label{fig:mnist3}
		\hspace{-8mm}
	}
	\subfloat[MNIST-MLP, FlexCFL-$\eta_g$]{
		\includegraphics{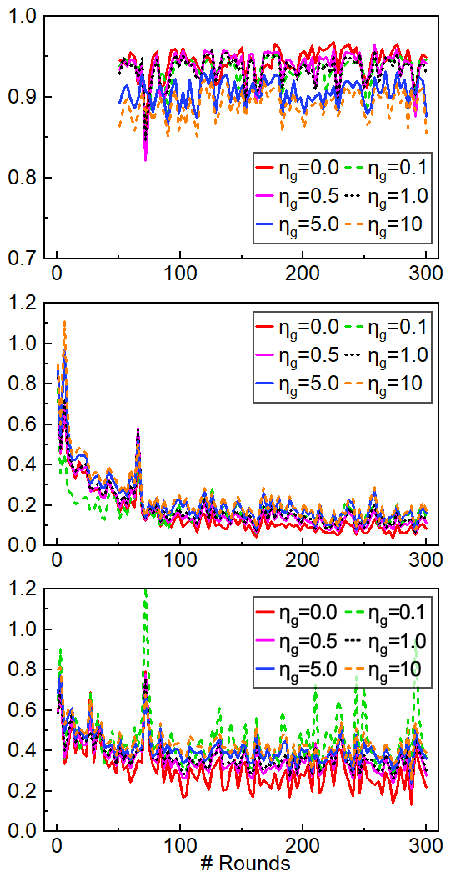}
		\label{fig:mnistlr}
		\hspace{-8mm}
	}
	\caption{Evaluation results on MNIST ($m=3$). Top: test accuracy; Middle: weighted training loss based on $K$ selected clients; Bottom: discrepancy between selected clients and server (FedAvg and FedProx) or weighted discrepancy between selected clients and groups (FlexCFL, FlexCFL-$\eta_g$ IFCA, FeSEM). The inter-group learning rate $\eta_g=0.1$ in (a), (b), (c).}
	\label{fig:acc1}
	\vspace{-4mm}
\end{figure*}

\begin{table*}[t]
	\scriptsize
	\centering
	\caption{Comparisons with FedAvg\cite{mcmahan2017communication}, IFCA\cite{ghosh2020efficient}, FeSEM\cite{xie2020multi}, FedGroup~\cite{duan2020fedgroup}, FlexCFL, FlexCFL with $\eta_g=5.0$ on MNIST, FEMNIST, Synthetic, FashionMNIST with three kinds of distribution shift (Swap all/Swap part/Incremental). The swap probability is 0.05 in swap all and swap part settings, 25\% of data is released every 50 rounds in incremental setting. Local Epoch $E=10$.}
	\begin{tabular}{|l|ccc|ccc|}
		\hline
		Dataset-Model & FedAvg & IFCA & FeSEM & FedGroup & FlexCFL & FlexCFL-$\eta_g$ \\ \hline
		MNIST-MCLR & $89.9/89.2/89.9$ & $64.5/85.0/90.2$ & $84.7/83.6/84.5$ & $87.1/86.3/95.1$ & $\bm{95.1}/\bm{91.7}/\bm{95.1}$ & $89.1/90.2/89.1$ \\
		MNIST-MLP & $92.0/92.7/92.3$ & $63.4/83.7/92.0$ & $89.6/89.3/84.0$ & $91.7/89.9/95.6$ & $\bm{93.8}/93.3/\bm{95.6}$ & $93.4/\bm{93.5}/92.4$ \\
		MNIST-CNN & $97.3/97.5/95.4$ & $66.5/86.6/92.6$ & $92.2/89.6/80.0$ & $96.5/95.7/97.3$ & $97.0/97.0/\bm{97.4}$ & $\bm{97.8}/\bm{97.9}/95.4$ \\
		FEMNIST-MCLR & $77.1/77.3/69.9$ & $76.3/56.7/77.3$ & $41.5/40.6/37.7$ & $58.1/64.6/85.8$ & $\bm{84.7}/68.5/84.8$ & $83.1/\bm{90.2}/\bm{89.1}$ \\
		FEMNIST-MLP & $80.3/\bm{82.1}/74.2$ & $78.7/63.4/78.7$ & $58.9/57.1/48.8$ & $66.1/67.8/87.5$ & $86.6/73.4/\bm{87.5}$ & $\bm{89.9}/81.5/85.7$ \\
		FEMNIST-CNN & $95.7/95.9/74.1$ & $86.1/84.0/82.8$ & $23.1/59.8/19.8$ & $87.6/90.9/88.2$ & $93.7/94.7/\bm{88.6}$ & $\bm{97.2}/\bm{96.6}/86.1$ \\
		Synthetic(1,1)-MCLR & $63.4/71.1/78.2$ & $\bm{89.3}/\bm{96.8}/\bm{91.6}$ & $57.4/77.3/69.4$ & $84.4/93.5/89.8$ & $85.0/91.7/89.8$ & $58.0/76.5/80.2$ \\
		FashionMNIST-CNN & $89.4/89.3/83.6$ & $73.8/82.2/85.6$ & $82.2/81.8/78.2$ & $86.0/86.4/\bm{88.8}$ & $90.8/87.5/88.7$ & $\bm{91.8}/\bm{89.3}/87.6$ \\ \hline
	\end{tabular}
	\label{tab:result2}
\end{table*}

\begin{figure*}[t]
	\centering
	\vspace{-8mm}
	\subfloat[MNIST-MCLR]{
		\includegraphics{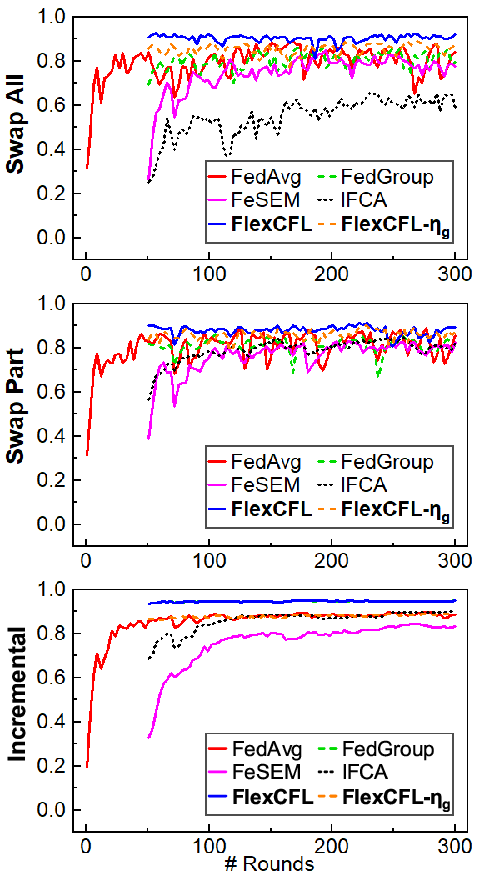}
		\label{fig:mnist_shift}
	}
	\subfloat[FEMNIST-CNN]{
		\includegraphics{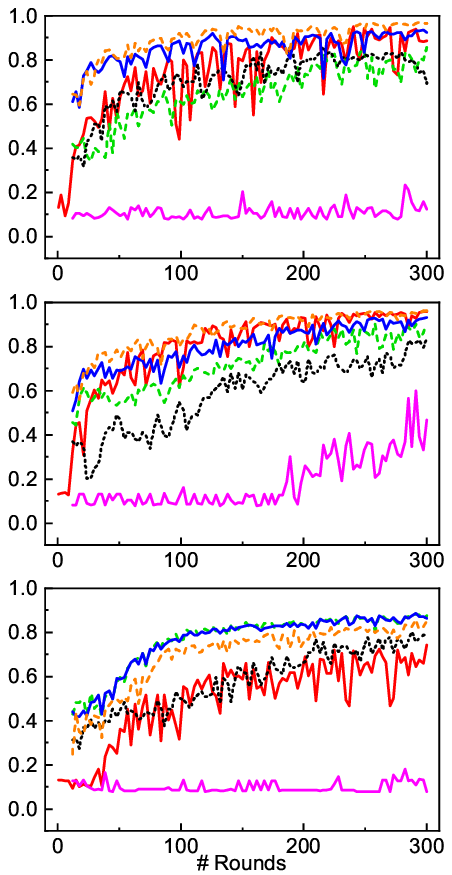}
		\label{fig:nist_shift}
	}
	\subfloat[FeshionMNIST-CNN]{
		\includegraphics{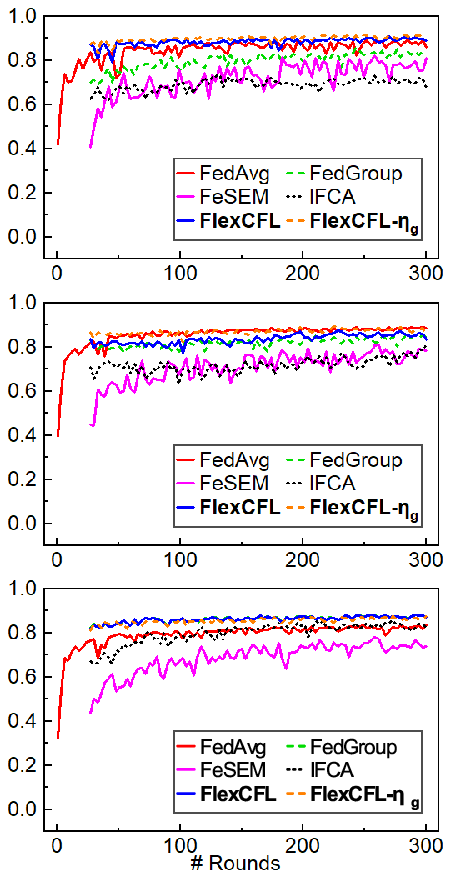}
		\label{fig:fashion_shift}
	}

	\caption{Test accuracy on MNIST ($m=3$), FEMNIST($m=5$), FashionMNIST($m=5$) with three kinds of distribution shift: Swap all (top); Swap part (middle); Incremental (bottom). The swap probability is 0.05 in the swap all and the swap part settings, 25\% of data is released every 50 rounds in the incremental setting. FedGroup is the static version of FlexCFL without client migration.}
	\label{fig:acc2}
	\vspace{-4mm}
\end{figure*}

\subsection{Effects of Proposed Framework}
\label{sec:effects}

We first compare the evaluated results of FlexCFL with baselines without involving distribution shift.
The results are shown in TABLE~\ref{tab:result1} and Fig.~\ref{fig:acc1}, we calculate the accuracy improvements related to the FeSEM.
The experiments show that FlexCFL, IFCA are significantly superior to other frameworks.
In particular, FlexCFL improves absolute test accuracy by $+8.4\%$ on MNIST, $+40.9\%$ on FEMNIST, $+11.3\%$ on FashionMNIST.

The FeSEM performs worst in all datasets compared to IFCA and FlexCFL, which can be interpreted as its failed client scheduling strategy.
For example, FeSEM clusters all clients into a group in MNIST-MLP, so these experimental results of FeSEM behave similarly to FedAvg (downgrade to one optimization direction).
The accuracy and discrepancy curves of FedProx illustrate that adding the proximal term can reduce the divergence caused by the heterogeneous data and make the training more stable (i.e. The green lines in Fig.~\subref*{fig:mnist1} shown that the accuracy of FedProx is most stable and the discrepancy is lowest).
However, there is not enough evidence to suggest that adding the proximal term is significantly helpful in improving accuracy and convergence speed.
An interesting observation is that although the discrepancy of FedProx is relatively low, its training loss is highest. On the other hand, the discrepancy of FlexCFL is higher than FedProx but can get higher accuracy. This means that bounding model discrepancy does not necessarily result in better performance when the incongruent optimization goals are not untangled.

The accuracy of IFCA is similar to (average $-1.33\%$ on MNIST, $-3.3\%$ on FEMNIST) that of FlexCFL, but FlexCFL has lower communication overhead in design, we will explain it later.
FlexCFL also shows a significant improvement in convergence speed compared to IFCA and FeSEM as shown in the training losses of them, which is helpful to reduce the communication consumption of the FL systems.

To investigate the effects of our proposed strategies, we further perform two ablation studies: RCC (random cluster centers) and RAC (randomly assign cold clients, but the clustered clients are retained in their groups).
In the RCC setting, the accuracy is moderately degraded (except FEMNIST-MLP and Synthetic) but still surpasses the FeSEM.
An implicit reason is that the client's data is randomly divided without any particular preference, so the random is a good estimate of the clustering center.	
The RAC strategy leads to a significant decrease in accuracy (average $-5.5\%$ on MNIST, $-19.7\%$ on FEMNIST) and the final scores are even worse than the FedAvg.
Therefore, the combination of our clustering algorithm and newcomer cold start strategy is efficient and can reach more improvements.

To explore the potential of semi-pluralistic architecture, we evaluate FlexCFL under difference $\eta_g$. 
The details are presented in Fig.~\subref*{fig:mnistlr}.
Our experiments in MNIST-MLP show that the inter-group aggregation mechanism with a proper learning rate (i.e. $\eta_g=0.1$) can slightly improve the convergence rate of model training.
Unfortunately, the convergence rates of other experimental sets do not improve as expected, which can be interpreted as each group in FlexCFL is highly specialized and has few common representations. 
However, in the distribution shift situation, our inter-group model aggregation mechanism can achieve advantages as shown in the next section.




\subsection{Distribution Shift}\label{sec:shift}
Some common types of distribution shift include covariance shift, label shift, concept shift~\cite{pan2009survey}. 
How to correct these distribution shifts is an open challenge in ML and beyond the scope of this work.
Therefore, we simulate three kinds of client-level distribution shift under CFL scenarios:
1) Swap all, we swap the local data (including training set and test set) of two random clients with a preset probability in each round of training;
2) Swap part, we swap two unique labels of the local data of two random clients with a preset probability in each round of training;
3) Incremental, we gradually release the training data of clients when the training reaches preset rounds.
It is worth noting that the above kinds of distribution shift will not change the global data distribution (union of client data), so we call it the client-level distribution shift.

We change the data distribution before the client scheduling process in each round, the experimental results are shown in TABLE~\ref{tab:result2} and Fig.~\ref{fig:acc2}. 
It can be concluded from Fig.~\ref{fig:acc2} that the distribution shifts have a negative impact on the training of the CFL-based frameworks, but the impact on FedAvg is negligible.
For example, although IFCA is promising in the no shift setting, the accuracy of IFCA, FeSEM, FedGroup on FEMNIST decreased by $-10.2\%$, $-11.8\%$, $-23.3\%$ in the swap all setting, and the accuracy gap between FedAvg in no shift and FedAvg in swap all is only $1.1\%$.
The main reason for this accuracy degradation is that our client-level distribution shift will change the distribution of client data, which will lead to some mismatches in clustered federated training. 
Since FedAvg trains the consensus model based on all clients and the global distribution is unchanged, the impact of client-level distribution shift on FedAvg is minimal.
The performance of FeSEM is still unsatisfactory and Fig.~\subref*{fig:nist_shift} shows that it cannot converge in FEMNIST-CNN.
FedGroup is the static version of FlexCFL and it also suffers from performance degradation.
Specifically, the absolute accuracy dropped by $-1.96\%$ on MNIST in swap all, $-19.4\%$ on FEMNIST in swap part, $-3.8\%$ on FashionMNIST.
FLexCFL can leverage the client migration strategy to correct the mismatches caused by distribution shift, the results show that it improves absolute test accuracy by $+2.2\%$ on MNIST, $+3.9\%$ on FEMNIST, $+1.4\%$ on FashionMNIST compared to FedAvg in the swap all setting.
For the swap part, FlexCFL improves by $+8.9\%$ on MNIST, $+10.8\%$ on FEMNIST, $+5.3\%$ on FashionMNIST compared to IFCA.
The evaluation results of FedGroup and FlexCFL in the incremental setting are almost identical ($+0.6\%$ in total) because the random release data policy has little effect on the client data distribution.

\begin{figure}[t]
	\centering
	\hspace{-3mm}
	\subfloat[FEMNIST-MLP; Accuracy]{
	\includegraphics{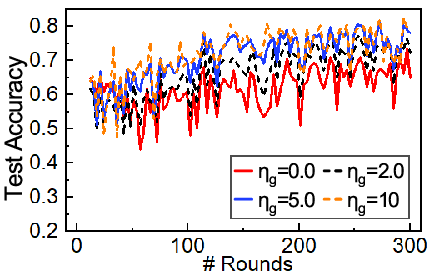}
}
	\hspace{-4mm}
	\subfloat[FEMNIST-MLP; Training loss]{
	\includegraphics{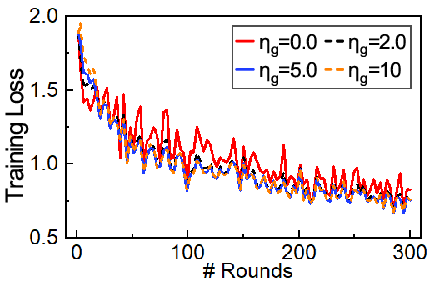}
}
\caption{Evaluation results of FlexCFL with different inter-group learning rate $\eta_g$ on FEMNIST-MLP in swap part setting.}\label{fig:nistlr_mlp_part_shift}
\end{figure}

Our inter-group aggregation strategy with $\eta_g=5.0$ shows a prominent advantage in the distribution shift setting as shown by FlexCFL-$\eta_g$ in the TABLE~\ref{tab:result2}.
Especially in the swap part setting, the accuracy increased by $+10.5\%$ on FEMNIST compared to FlexCFL without inter-group aggregation.
We further study the impact of difference $\eta_g$ in Fig.~\ref{fig:nistlr_mlp_part_shift}.
The results show that the higher inter-group learning rate can improve model accuracy and make the training more stable, which is very different from the previous results in Fig.\subref*{fig:mnistlr}.
This is because in the swap part or swap all setting, there is a data exchange between clients, which leads to the assimilation of group optimization goals, so the group model can gain advantages from model sharing.
Therefore, the above advantages are not clearly observed in the incremental setting.

\begin{figure}
	\centering
	\vspace{-3mm}
	\subfloat[MNIST-MCLR; Swap all]{
	\includegraphics{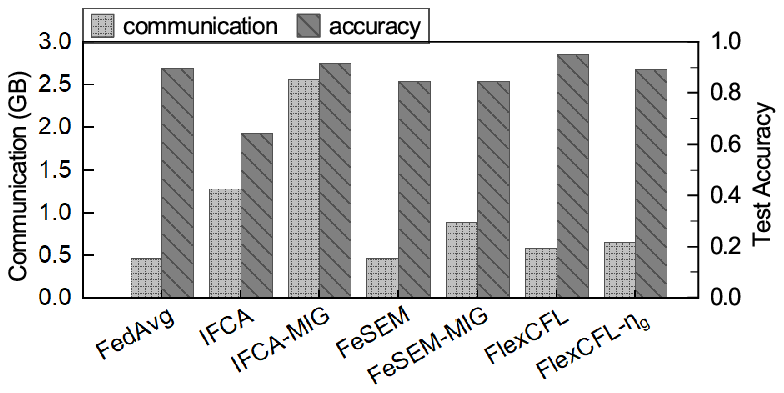}
	\label{fig:overhead_mnist}
}\\
\vspace{-3mm}
	\subfloat[FEMNIST-MLP; Swap part]{
	\includegraphics{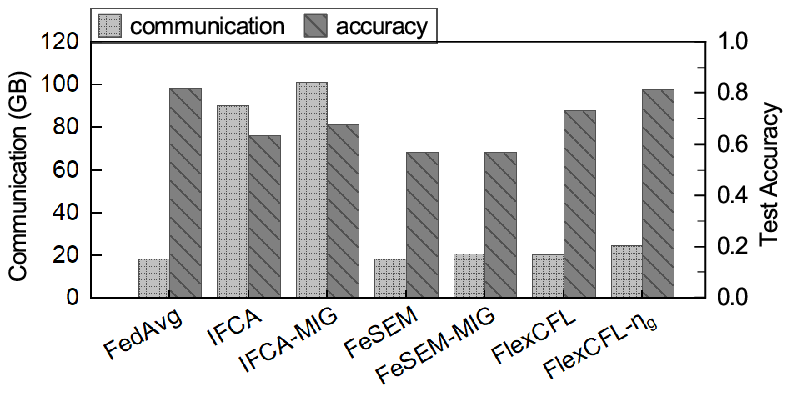}
	\label{fig:overhead_nist}
}\\
\vspace{-3mm}
	\subfloat[FashionMNIST; Incremental]{
	\includegraphics{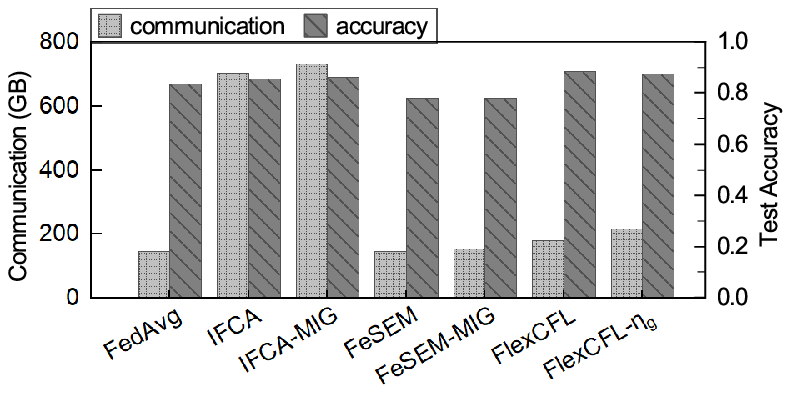}
}
	\caption{Communication consumption of FL frameworks train 300 rounds in the client-level distribution shift setting. IFCA-MIG and FeSEM-MIG are the revisions with the client migration strategy. $\eta_g=5.0$}
	\label{fig:overhead}
\end{figure}

Of course, to handle the client-level distribution shift issue, IFCA and FeSEM can use the client migration strategy like FlexCFL. However, due to the runtime clustering algorithms of IFCA, clients need to download all group models when migration is required, such improvement means huge additional communication overhead. We compare the communication consumption of different frameworks in Fig.~\ref{fig:overhead}. We denote IFCA with client migration as IFCA-MIG, FeSEM with client migration as FeSEM-MIG.
First of all, since the scheduling criterion of FeSEM is based on the $\ell_2$ distance between the local model and group model and has nothing to do with local data, so FeSEM and FeSEM-MIG have the same accuracy.
In addition, although the communication requirement of FeSEM is as low as FedAvg, its accuracy is miserable.
IFCA-MIG shows satisfactory accuracy, but its communication overhead is expensive.
For example, in order to get $+1.9\%$ accuracy improvement, the communication requirement of IFCA-MIG is $\times 5.48$ that of FedAvg  (Fig.~\subref*{fig:overhead_mnist}). 
In contrast, FlexCFL is more efficient in communication, $\times 1.26$ communication consumption can achieve $5.2\%$ accuracy improvement compared to FedAvg.
As shown in Fig.~\subref*{fig:overhead_nist}, FlexCFL-$\eta_g$ can further improve $+8.1\%$ accuracy with additional $11.4\%$ communication compared to FlexCFL.
Moreover, the convergence speed of FlexCFL is faster than FedAvg, so the number of communication rounds to reach the target accuracy is smaller, which means less communication is actually required.

\section{Conclusion}\label{sec:conclusion}
In this work, we have presented a flexible clustered federated learning frameworks FlexCFL, which can improve the model performance of federated training by efficient clustering and cold start strategies.
We evaluated the proposed frameworks on four open datasets and shown the superiority of FlexCFL compared to FedAvg, FedProx, FeSEM.
FlexCFL significantly improved $+40.9\%$ top-1 test accuracy on FEMNIST compared to FedSEM.
Our evaluations on 8 models shown that FlexCFL achieved higher classification accuracy compared to FedAvg, FedProx, FeSEM and the random strategy RAC.
Besides, we have found that FlexCFL can strike a balance between communication and accuracy in the client-level distribution shift environment.

\ifCLASSOPTIONcaptionsoff
  \newpage
\fi



\bibliographystyle{IEEEtran}
\balance
\bibliography{IEEEabrv,references}

\begin{thebibliography}{10}
\providecommand{\url}[1]{#1}
\csname url@samestyle\endcsname
\providecommand{\newblock}{\relax}
\providecommand{\bibinfo}[2]{#2}
\providecommand{\BIBentrySTDinterwordspacing}{\spaceskip=0pt\relax}
\providecommand{\BIBentryALTinterwordstretchfactor}{4}
\providecommand{\BIBentryALTinterwordspacing}{\spaceskip=\fontdimen2\font plus
\BIBentryALTinterwordstretchfactor\fontdimen3\font minus
  \fontdimen4\font\relax}
\providecommand{\BIBforeignlanguage}[2]{{%
\expandafter\ifx\csname l@#1\endcsname\relax
\typeout{** WARNING: IEEEtran.bst: No hyphenation pattern has been}%
\typeout{** loaded for the language `#1'. Using the pattern for}%
\typeout{** the default language instead.}%
\else
\language=\csname l@#1\endcsname
\fi
#2}}
\providecommand{\BIBdecl}{\relax}
\BIBdecl

\bibitem{duan2020fedgroup}
M.~Duan, D.~Liu, X.~Ji, R.~Liu, L.~Liang, X.~Chen, and Y.~Tan, ``Fedgroup:
  Efficient clustered federated learning via decomposed data-driven measure,''
  \emph{arXiv preprint arXiv:2010.06870}, 2020.

\bibitem{konevcny2016federated}
J.~Kone{\v{c}}n{\`y}, H.~B. McMahan, F.~X. Yu, P.~Richt{\'a}rik, A.~T. Suresh,
  and D.~Bacon, ``Federated learning: Strategies for improving communication
  efficiency,'' \emph{arXiv preprint arXiv:1610.05492}, 2016.

\bibitem{mcmahan2017communication}
B.~McMahan, E.~Moore, D.~Ramage, S.~Hampson, and B.~A. y~Arcas,
  ``Communication-efficient learning of deep networks from decentralized
  data,'' in \emph{Proceedings of the 20th International Conference on
  Artificial Intelligence and Statistics (AISTATS)}, 2017, pp. 1273--1282.

\bibitem{bonawitz2019towards}
K.~Bonawitz, H.~Eichner, W.~Grieskamp, D.~Huba, A.~Ingerman, V.~Ivanov,
  C.~Kiddon, J.~Konecny, S.~Mazzocchi, H.~B. McMahan \emph{et~al.}, ``Towards
  federated learning at scale: System design,'' in \emph{Proceedings of the 2nd
  SysML Conference}, 2019.

\bibitem{yang2019federated}
Q.~Yang, Y.~Liu, T.~Chen, and Y.~Tong, ``Federated machine learning: Concept
  and applications,'' \emph{ACM Transactions on Intelligent Systems and
  Technology (TIST)}, vol.~10, no.~2, pp. 1--19, 2019.

\bibitem{li2020federated}
T.~Li, A.~K. Sahu, A.~Talwalkar, and V.~Smith, ``Federated learning:
  Challenges, methods, and future directions,'' \emph{IEEE Signal Processing
  Magazine}, vol.~37, no.~3, pp. 50--60, 2020.

\bibitem{li2014scaling}
M.~Li, D.~G. Andersen, J.~W. Park, A.~J. Smola, A.~Ahmed, V.~Josifovski,
  J.~Long, E.~J. Shekita, and B.-Y. Su, ``Scaling distributed machine learning
  with the parameter server,'' in \emph{Proceedings of the 11th USENIX
  Symposium on Operating Systems Design and Implementation (OSDI)}, 2014, pp.
  583--598.

\bibitem{zhao2018federated}
Y.~Zhao, M.~Li, L.~Lai, N.~Suda, D.~Civin, and V.~Chandra, ``Federated learning
  with non-iid data,'' \emph{arXiv preprint arXiv:1806.00582}, 2018.

\bibitem{krizhevsky2009learning}
A.~Krizhevsky and G.~Hinton, ``Learning multiple layers of features from tiny
  images,'' Citeseer, Tech. Rep., 2009.

\bibitem{Simonyan15}
K.~Simonyan and A.~Zisserman, ``Very deep convolutional networks for
  large-scale image recognition,'' in \emph{Proceedings of the 3rd
  International Conference on Learning Representations (ICLR)}.\hskip 1em plus
  0.5em minus 0.4em\relax IEEE, 2015.

\bibitem{sattler2019robust}
F.~Sattler, S.~Wiedemann, K.-R. M{\"u}ller, and W.~Samek, ``Robust and
  communication-efficient federated learning from {non-i.i.d.} data,''
  \emph{IEEE Transactions on Neural Networks and Learning Systems (TNNLS)}, pp.
  1--14, 2019.

\bibitem{Li2020On}
X.~Li, K.~Huang, W.~Yang, S.~Wang, and Z.~Zhang, ``On the convergence of fedavg
  on non-iid data,'' in \emph{Proceedings of the 8th International Conference
  on Learning Representations (ICLR)}, 2020.

\bibitem{duan2019astraea}
M.~Duan, D.~Liu, X.~Chen, Y.~Tan, J.~Ren, L.~Qiao, and L.~Liang, ``Astraea:
  Self-balancing federated learning for improving classification accuracy of
  mobile deep learning applications,'' in \emph{Proceedings of the IEEE 37th
  International Conference on Computer Design (ICCD)}.\hskip 1em plus 0.5em
  minus 0.4em\relax IEEE, 2019, pp. 246--254.

\bibitem{sattler2020clustered}
F.~Sattler, K.-R. M{\"u}ller, and W.~Samek, ``Clustered federated learning:
  Model-agnostic distributed multitask optimization under privacy
  constraints,'' \emph{IEEE Transactions on Neural Networks and Learning
  Systems (TNNLS)}, pp. 1--13, 2020.

\bibitem{sattler2020byzantine}
F.~Sattler, K.-R. M{\"u}ller, T.~Wiegand, and W.~Samek, ``On the byzantine
  robustness of clustered federated learning,'' in \emph{Proceedings of the
  IEEE International Conference on Acoustics, Speech and Signal Processing
  (ICASSP)}.\hskip 1em plus 0.5em minus 0.4em\relax IEEE, 2020, pp. 8861--8865.

\bibitem{xie2020multi}
M.~Xie, G.~Long, T.~Shen, T.~Zhou, X.~Wang, and J.~Jiang, ``Multi-center
  federated learning,'' \emph{arXiv preprint arXiv:2005.01026}, 2020.

\bibitem{ghosh2020efficient}
A.~Ghosh, J.~Chung, D.~Yin, and K.~Ramchandran, ``An efficient framework for
  clustered federated learning,'' in \emph{Advances in Neural Information
  Processing Systems}, vol.~33.\hskip 1em plus 0.5em minus 0.4em\relax Curran
  Associates, Inc., 2020, pp. 19\,586--19\,597.

\bibitem{briggs2020federated}
C.~{Briggs}, Z.~{Fan}, and P.~{Andras}, ``Federated learning with hierarchical
  clustering of local updates to improve training on non-{IID} data,'' in
  \emph{Proceedings of the IEEE International Joint Conference on Neural
  Networks (IJCNN)}, 2020, pp. 1--9.

\bibitem{sarkar2019perfect}
S.~Sarkar and A.~K. Ghosh, ``On perfect clustering of high dimension, low
  sample size data,'' \emph{IEEE transactions on pattern analysis and machine
  intelligence (TPAMI)}, vol.~42, no.~9, pp. 2257--2272, 2019.

\bibitem{lecun1998gradient}
Y.~LeCun, L.~Bottou, Y.~Bengio, and P.~Haffner, ``Gradient-based learning
  applied to document recognition,'' \emph{Proceedings of the IEEE}, vol.~86,
  no.~11, pp. 2278--2324, 1998.

\bibitem{cohen2017emnist}
G.~Cohen, S.~Afshar, J.~Tapson, and A.~van Schaik, ``{EMNIST}: Extending mnist
  to handwritten letters,'' in \emph{Proceedings of the 2017 International
  Joint Conference on Neural Networks (IJCNN)}.\hskip 1em plus 0.5em minus
  0.4em\relax IEEE, 2017, pp. 2921--2926.

\bibitem{xiao2017fashion}
H.~Xiao, K.~Rasul, and R.~Vollgraf, ``Fashion-mnist: a novel image dataset for
  benchmarking machine learning algorithms,'' \emph{arXiv preprint
  arXiv:1708.07747}, 2017.

\bibitem{li2018federated}
T.~Li, A.~K. Sahu, M.~Zaheer, M.~Sanjabi, A.~Talwalkar, and V.~Smith,
  ``Federated optimization in heterogeneous networks,'' in \emph{Proceedings of
  the 3rd SysML Conference}, 2020.

\bibitem{duan2020self}
M.~Duan, D.~Liu, X.~Chen, R.~Liu, Y.~Tan, and L.~Liang, ``Self-balancing
  federated learning with global imbalanced data in mobile systems,''
  \emph{IEEE Transactions on Parallel and Distributed Systems (TPDS)}, vol.~32,
  no.~1, pp. 59--71, 2020.

\bibitem{wang2021addressing}
L.~Wang, S.~Xu, X.~Wang, and Q.~Zhu, ``Addressing class imbalance in federated
  learning,'' in \emph{Proceedings of the 35th AAAI Conference on Artificial
  Intelligence (AAAI)}, vol.~35, no.~11, 2021, pp. 10\,165--10\,173.

\bibitem{lee2020tornadoaggregate}
J.-w. Lee, J.~Oh, S.~Lim, S.-Y. Yun, and J.-G. Lee, ``Tornadoaggregate:
  Accurate and scalable federated learning via the ring-based architecture,''
  \emph{arXiv preprint arXiv:2012.03214}, 2020.

\bibitem{radovanovic2010existence}
M.~Radovanovi{\'c}, A.~Nanopoulos, and M.~Ivanovi{\'c}, ``On the existence of
  obstinate results in vector space models,'' in \emph{Proceedings of the 33rd
  international ACM SIGIR conference on Research and development in information
  retrieval}, 2010, pp. 186--193.

\bibitem{dwork2015reusable}
C.~Dwork, V.~Feldman, M.~Hardt, T.~Pitassi, O.~Reingold, and A.~Roth, ``The
  reusable holdout: Preserving validity in adaptive data analysis,''
  \emph{Science}, vol. 349, no. 6248, pp. 636--638, 2015.

\bibitem{liu2020client}
L.~Liu, J.~Zhang, S.~Song, and K.~B. Letaief, ``Client-edge-cloud hierarchical
  federated learning,'' in \emph{Proceedings of the IEEE International
  Conference on Communications (ICC)}.\hskip 1em plus 0.5em minus 0.4em\relax
  IEEE, 2020, pp. 1--6.

\bibitem{abdulrahman2020fedmccs}
S.~AbdulRahman, H.~Tout, A.~Mourad, and C.~Talhi, ``Fedmccs: multicriteria
  client selection model for optimal iot federated learning,'' \emph{IEEE
  Internet of Things Journal}, vol.~8, no.~6, pp. 4723--4735, 2020.

\bibitem{smith2017federated}
V.~Smith, C.-K. Chiang, M.~Sanjabi, and A.~Talwalkar, ``Federated multi-task
  learning,'' in \emph{Proceedings of the 31st International Conference on
  Neural Information Processing Systems (NeurIPS)}, 2017, pp. 4427--4437.

\bibitem{bonawitz2017practical}
K.~Bonawitz, V.~Ivanov, B.~Kreuter, A.~Marcedone, H.~B. McMahan, S.~Patel,
  D.~Ramage, A.~Segal, and K.~Seth, ``Practical secure aggregation for
  privacy-preserving machine learning,'' in \emph{Proceedings of the 2017 ACM
  SIGSAC Conference on Computer and Communications Security (CCS)}.\hskip 1em
  plus 0.5em minus 0.4em\relax ACM, 2017, pp. 1175--1191.

\bibitem{mohri2019agnostic}
M.~Mohri, G.~Sivek, and A.~T. Suresh, ``Agnostic federated learning,'' in
  \emph{Proceedings of the 36th International Conference on Machine Learning
  (ICML)}, 2019, pp. 4615--4625.

\bibitem{golub1971singular}
G.~H. Golub and C.~Reinsch, ``Singular value decomposition and least squares
  solutions,'' in \emph{Linear Algebra}.\hskip 1em plus 0.5em minus 0.4em\relax
  Springer, 1971, pp. 134--151.

\bibitem{arthur2006k}
D.~Arthur and S.~Vassilvitskii, ``k-means++: The advantages of careful
  seeding,'' Stanford, Tech. Rep., 2006.

\bibitem{miller2020effect}
J.~Miller, K.~Krauth, B.~Recht, and L.~Schmidt, ``The effect of natural
  distribution shift on question answering models,'' in \emph{Proceedings of
  the 37th International Conference on Machine Learning (ICML)}.\hskip 1em plus
  0.5em minus 0.4em\relax PMLR, 2020, pp. 6905--6916.

\bibitem{wang2019adaptive}
S.~Wang, T.~Tuor, T.~Salonidis, K.~K. Leung, C.~Makaya, T.~He, and K.~Chan,
  ``Adaptive federated learning in resource constrained edge computing
  systems,'' \emph{IEEE Journal on Selected Areas in Communications}, vol.~37,
  no.~6, pp. 1205--1221, 2019.

\bibitem{stich2018local}
S.~U. Stich, ``Local {SGD} converges fast and communicates little,'' in
  \emph{Proceedings of the 7th International Conference on Learning
  Representations (ICLR)}, 2019.

\bibitem{abadi2016tensorflow}
M.~Abadi, P.~Barham, J.~Chen, Z.~Chen, A.~Davis, J.~Dean, M.~Devin,
  S.~Ghemawat, G.~Irving, M.~Isard \emph{et~al.}, ``Tensorflow: A system for
  large-scale machine learning,'' in \emph{Proceedings of the 12th USENIX
  Symposium on Operating Systems Design and Implementation (OSDI)}, 2016, pp.
  265--283.

\bibitem{shamir2014communication}
O.~Shamir, N.~Srebro, and T.~Zhang, ``Communication-efficient distributed
  optimization using an approximate newton-type method,'' in \emph{Proceedings
  of the 31th International conference on machine learning (ICML)}, 2014, pp.
  1000--1008.

\bibitem{li2020ditto}
T.~Li, S.~Hu, A.~Beirami, and V.~Smith, ``Ditto: Fair and robust federated
  learning through personalization,'' \emph{arXiv preprint arXiv:2012.04221},
  2020.

\bibitem{yao2007early}
Y.~Yao, L.~Rosasco, and A.~Caponnetto, ``On early stopping in gradient descent
  learning,'' \emph{Constructive Approximation}, vol.~26, no.~2, pp. 289--315,
  2007.

\bibitem{pan2009survey}
S.~J. Pan and Q.~Yang, ``A survey on transfer learning,'' \emph{IEEE
  Transactions on knowledge and data engineering (TKDE)}, vol.~22, no.~10, pp.
  1345--1359, 2009.

\end{thebibliography}
\end{document}